%% file: main.tex
\documentclass{article}
\PassOptionsToPackage{numbers,sort&compress}{natbib}
\usepackage[preprint]{neurips_2026}
\usepackage[T1]{fontenc}
\usepackage[utf8]{inputenc}
\usepackage{microtype}
\usepackage{amsmath,amssymb}
\usepackage{graphicx,booktabs,longtable,array}
\newcolumntype{P}[1]{>{\raggedright\arraybackslash}p{#1}}
\usepackage{xcolor}
\usepackage[colorlinks=true,allcolors=blue!45!black]{hyperref}
\graphicspath{{figures/}}
\makeatletter
\renewcommand{\@noticestring}{preprint}
\makeatother

\title{When gene summaries reveal the answers: Gene Ontology circularity in gene embedding benchmarks}
\author{%
  Jiawei Wu\\
  Rutgers University\\
  \href{mailto:jw2070@rutgers.edu}{\texttt{jw2070@rutgers.edu}}
  \And
  Huishu Wang\\
  New York University\\
  \href{mailto:hw2804@nyu.edu}{\texttt{hw2804@nyu.edu}}
  \AND
  Junlin Chen\\
  Rutgers University\\
  \href{mailto:junlin.chen110@rutgers.edu}{\texttt{junlin.chen110@rutgers.edu}}
  \And
  Hongyi Wang\thanks{Corresponding author.}\\
  Rutgers University\\
  \href{mailto:hw689@cs.rutgers.edu}{\texttt{hw689@cs.rutgers.edu}}
}
\hypersetup{
  pdftitle={When gene summaries reveal the answers: Gene Ontology circularity in gene embedding benchmarks},
  pdfauthor={Jiawei Wu, Huishu Wang, Junlin Chen, Hongyi Wang}
}

\begin{document}
\maketitle

\begin{abstract}
A gene summary can contain the very annotations used to evaluate an embedding of its text.
The Alliance of Genome Resources converts Gene Ontology (GO) annotations into sentences, which
NCBI Gene displays and GenePT includes in its input. These summaries covered 6,172 of
19,256 human protein-coding genes (32.1\%), including 58\% of genes with one to ten linked
publications. We embedded the same corpus with and without the generated summaries and refitted
a function-prediction probe on each version. Among genes with one to ten publications, removal
reduced the reported per-gene area under the receiver operating characteristic curve (AUROC) by 0.078 with a linear probe (95\% bootstrap interval
0.072--0.084) and by 0.033 with a multilayer perceptron. These estimates use a historical GO
scorer that mishandled ties. The original predictions are unavailable for rescoring.
A simple TF-IDF encoder also benefited from the summaries, as did models evaluated only on
experimental annotations. Replacing generated summaries with UniProt function descriptions
recovered much of the lost performance despite little verbatim overlap. UniProt also curates
many of the evaluation labels. In mouse and yeast, we found a similar dependence on the source
of the description. We introduce GeneProv to compare encoders on matched text and to examine
how representation rankings change across GO recovery, non-GO endpoints and co-essentiality.
\end{abstract}

\begin{center}
  \small\textbf{GeneProv benchmark:} \url{https://hwang595.github.io/gene-embedding-exposure/}
\end{center}

\section{Introduction}

A gene summary may tell a reader where a protein is located. If a model reads that summary before
predicting the protein's location, its score depends in part on where the sentence came from. GenePT embeds NCBI Gene summaries and performs well on
gene-level tasks \citep{chen2024genept,brown2015ncbigene}. Other methods use gene descriptions
to represent cells or predict perturbation responses \citep{liu2023scelmo,istrate2024scgenept}.
In comparisons with sequence, expression and network representations, text embeddings often
rank near the top for gene-function prediction
\citep{kantor2024genebenchmark,zhong2025benchmark,huang2026gebench}. The interpretation of
those rankings depends on who wrote the summary and what information they used.

Since February 2022, NCBI Gene has filled gaps in RefSeq coverage with descriptions from the
Alliance of Genome Resources \citep{ncbi2022insights,oleary2016refseq}. The Alliance generates
sentences from a gene's GO annotations \citep{kishore2020alliance,alliance2024}. A model that
encodes these sentences and is evaluated on GO has therefore received some of its targets in
prose. The genes most likely to lack a curator-written summary are also of particular interest
in benchmarks for understudied genes.

Annotation circularity and uneven literature coverage have long complicated gene-function
prediction \citep{gillis2013,brechtmann2023,jararweh2024litgene}. SPINDOCTOR explicitly discussed
the possibility of recovering GO annotations from generated descriptions \citep{joachimiak2023spindoctor}.
DEGEmbedR distinguishes automated Alliance descriptions from other sources \citep{tan2026degembedr},
while GeneTEA and LitGene deliberately combine text and functional annotations
\citep{boyle2025genetea,macaulay2026litgene}. GenePT checked whether summaries named interacting gene
pairs \citep[App.~B.3]{chen2023geneptbiorxiv}. Zhong et al.\ restricted GO evaluation to experimental
evidence and to annotations added after a cutoff \citep{zhong2025benchmark}. An experimental result
can still appear in a generated sentence, and even an older description can supply a gene's
prior annotation profile.

We ask how often these descriptions occur, how much they contribute to reported benchmark
scores, and which comparisons change when they are removed. We trace their provenance in two
public text archives, vary the input while holding the evaluation genes and folds fixed, and
repeat the comparison in published benchmark protocols. Replacements with UniProt prose or
LLM-written descriptions help distinguish direct generation from other ways that text can
share a source with a label. GeneProv extends the comparison to tasks with different sources
of evidence. Its fixed-text track measures the effect of changing the corpus, and its open
track compares representations with their inputs documented.

\section{Results}

\subsection{A third of human protein-coding genes carry GO-generated NCBI summaries}\label{sec:prev}

\begin{figure}[tp]
  \centering
  \includegraphics[width=\textwidth]{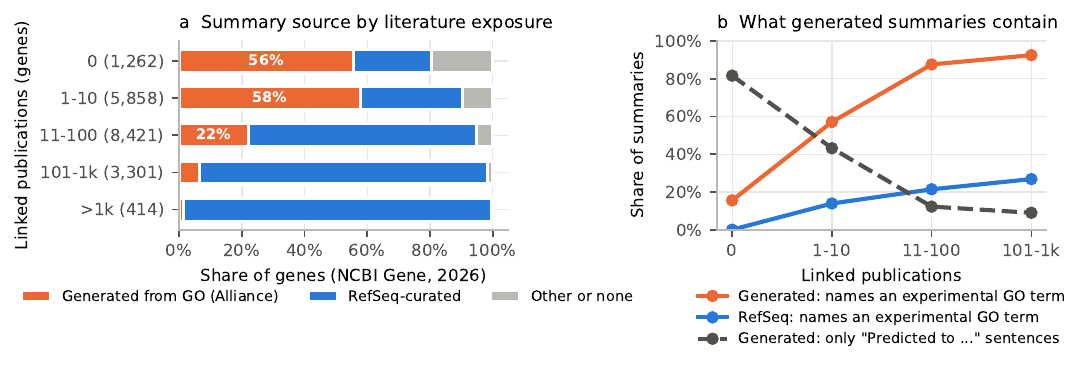}

  \caption{\textbf{GO-derived summaries are concentrated among genes with few linked publications.}
  (a) Source of the NCBI Gene summary by literature exposure in the human protein-coding universe, October 2026;
  numbers of genes are in brackets. (b) Share of generated (orange) and RefSeq (blue) summaries naming at least one
  experimentally supported GO term annotated to the gene, and share of generated summaries consisting only of
  ``Predicted to'' sentences (grey).}

  \label{fig:prevalence}
\end{figure}

On 6 October 2026, 6,172 of the 19,256 human protein-coding genes in our study (32.1\%) had an
NCBI Gene summary tagged “provided by Alliance of Genome Resources”. The proportion varied sharply
with the number of publications linked to the gene. It was 56\% for genes with no linked publications,
58\% for those with 1--10, and 6\% for those with 101--1000 (Fig.~\ref{fig:prevalence}a). The Alliance
builds these descriptions from templates and gives preference to experimentally supported annotations
\citep{kishore2020alliance}. Among genes with 1--10 publications, 57\% of Alliance descriptions named
at least one experimentally supported GO term for the gene verbatim, compared with 14\% of RefSeq
descriptions. For genes with 11--100 publications, the proportions were 88\% and 21\%
(Fig.~\ref{fig:prevalence}b). Averaged over genes, generated summaries contained the exact wording of
29\% of benchmark terms, compared with 1.4\% for RefSeq summaries. A Disease Ontology annotation appeared
in 480 generated summaries, in a sentence beginning “Implicated in”.

Provenance was harder to trace in the text distributed with GenePT. Its October 2023 archive
\citep{genept2023zenodo} contained 6,182 Alliance tags dated April 2022. These mapped to 5,851 genes
in our study, of which 5,808 still had an Alliance summary in 2026. The 2024 release
\citep{genept2024zenodo}, which supplied the NCBI text for the released ada-002 embeddings, retained
the sentences and stripped the tags. We used the opening words of the summaries to identify their source.
Against the 2023 tags, the planned list of templates had recall 0.997. Adding one phrase after inspection
raised recall to 0.999 with precision 1.000. The revised detector also reached 0.999 recall and 1.000
precision on the 2026 summaries with their tags hidden. The snapshots overlap, so these figures describe
performance on the audited texts.

\subsection{Removing generated summaries has the largest effect at 1--10 publications}\label{sec:inflate}

We encoded GenePT's text with Qwen3-Embedding-8B \citep{zhang2025qwen3embedding}. In one version
we kept every summary. In the other we removed the generated summaries from 5,851 genes, leaving
“Gene Symbol X” for each. We fitted a separate probe to each corpus and evaluated both on the same
held-out genes. With a linear probe, average per-gene GO AUROC fell from 0.883 to 0.849
(Table~\ref{tab:exposure}). The largest difference was 0.078 (0.072--0.084), for genes with 1--10
publications. It was 0.027 for genes with none, 0.026 for those with 11--100, and small above 100
(Fig.~\ref{fig:leak}a,b). With a multilayer perceptron (MLP), the difference was 0.033 at 1--10
publications and 0.010 overall (Supplementary Table~\ref{tab:mlp}). These are the original out-of-fold
estimates. The historical GO scorer broke ties by rank order. Quantifying the effect of the correction
requires the original predictions (Methods).

Per-gene AUROC asks whether an annotated term ranks above an unannotated term for the same gene.
Per-term AUROC asks whether a gene carrying a given annotation ranks above a gene without it.
For the linear probe, removal reduced mean per-term AUROC from 0.825 to 0.814 overall and from
0.871 to 0.780 within the 1--10 stratum. The corresponding MLP values in that stratum were
0.908 and 0.832 (Supplementary Table~\ref{tab:perterm}).

For exposure stratum $B$, we define the removal contrast as

\begin{equation}
\Delta_B = \frac{1}{|E_B|}\sum_{g\in E_B}
  \left[\operatorname{AUC}_g\!\left(\widehat f_{A,-k(g)}(X^A_g)\right)
       -\operatorname{AUC}_g\!\left(\widehat f_{L,-k(g)}(X^L_g)\right)\right],
\end{equation}
where $E_B$ is the same set of evaluable genes in both arms, $A$ and $L$ are the as-found and summary-filtered
corpora, and $\widehat f_{c,-k(g)}$ is the probe fitted outside gene $g$'s fold using corpus $c$.
Changing the corpus affects both the training examples and the held-out inputs. Isolating the effect
of editing only a test gene's summary would require evaluating both versions with the same fitted
probe. The paired bootstrap interval conditions on the fits used here. It covers sampling variation
across evaluation genes.

Absolute scores are harder to interpret. An unannotated gene--term pair can represent a function
that has yet to be recorded. A predictor that assigns every gene the training frequency of each term
reaches 0.861 for genes without publications and 0.808 for those with 1--10
(Table~\ref{tab:exposure}, post hoc). We therefore compare inputs within each exposure stratum.
Across three additional fold seeds, the removal contrast at 1--10 publications ranged from 0.076 to
0.079 for the linear probe and from 0.031 to 0.033 for the MLP. We set aside the nearest-neighbour
probe (0.064, 0.058--0.070 for the same stratum). Its label-vector predictions produce many ties,
making this estimate particularly sensitive to the scoring error.

\subsection{Replacement controls distinguish direct label text from other gene descriptions}\label{sec:prov}

\begin{figure}[tp]
  \centering
  \includegraphics[width=\textwidth]{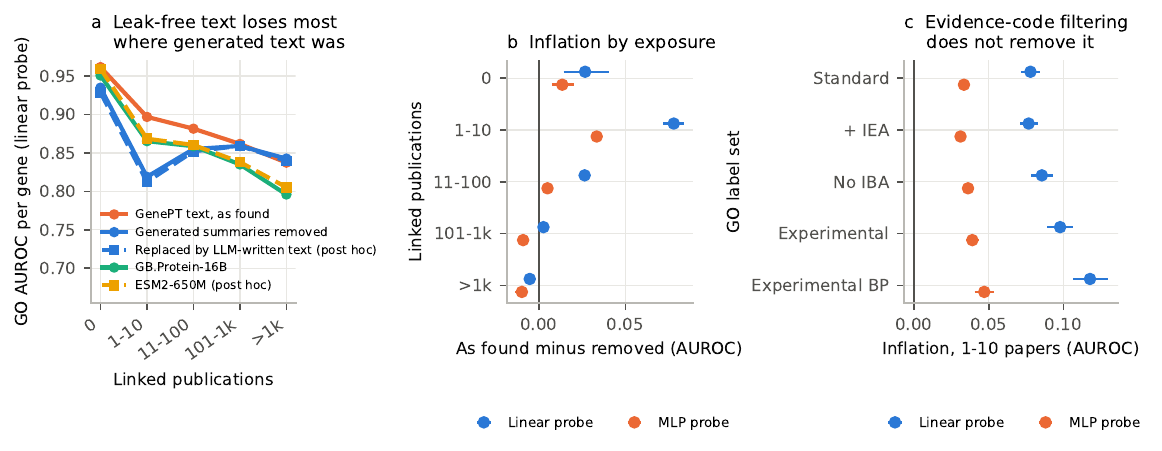}\\[2pt]
  \includegraphics[width=0.94\textwidth]{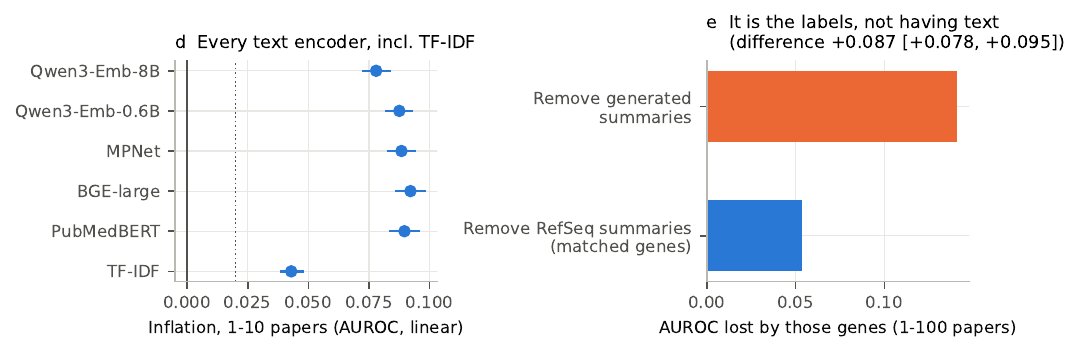}
\caption{\textbf{Effect of generated summaries on GO recovery.} (a) Per-gene AUROC by publication
  count for GenePT text, two replacements (symbol alone or LLM-written text), and two protein models.
  The label “leak-free” in the original plots denotes only removal of Alliance-generated summaries.
  The LLM and ESM2-650M comparisons were exploratory. (b) Paired difference between probes retrained
  on the original and filtered corpora, with 95\% bootstrap intervals. (c,d) The same difference for
  genes with 1--10 publications across five label sets (c) and six encoders (d). The dotted line marks
  the planned threshold of 0.02. (e) Effect of removing generated or RefSeq summaries among genes
  matched by exposure and GO-term count.}

  \label{fig:leak}
\end{figure}

How much of the loss comes from withholding any description of a gene? We matched genes with RefSeq
summaries to those with generated summaries by exposure and GO-term count. Removing generated
summaries reduced mean per-gene AUROC by 0.141 among the affected genes. Removing matched RefSeq
summaries reduced it by 0.054, a difference of 0.087 (0.078--0.095, Fig.~\ref{fig:leak}e).
In a post hoc control we replaced each generated summary with a description written by
Qwen2.5-7B-Instruct (Section~\ref{sec:llm}) and used the same encoder. Among genes with 1--10
publications, replacement cost 0.084 (0.078--0.091) relative to the original text and scored
0.007 (0.002--0.011) below the symbol alone. The corresponding MLP differences were 0.045 and
0.012. With experimental labels the loss relative to the original text was 0.107 (Fig.~\ref{fig:leak}a).
The replacements varied in length, accuracy and detail, each of which could contribute to these differences.

UniProt function descriptions gave a different result. The descriptions were comparable in length
to the generated summaries (median 278 and 213 characters, respectively) \citep{uniprot2025}.
We expected them to recover less than half the performance lost on removal. Of the 4,591 genes with
both texts, 1,938 evaluable genes had 1--10 publications. Their AUROC was 0.905 with the UniProt text against
0.909 with the generated summary, a difference of $+0.004$ in favour of the generated summary
(95\% interval $-0.004$ to $+0.012$). UniProt scored 0.149 above the symbol alone.
The difference between sources was also small with the MLP ($-0.002$) and in the 11--100 stratum
($+0.007$). For the 95 evaluable genes without publications, replacing or removing the generated text
each cost 0.061. Experimental labels gave the generated summary a modest advantage (0.020, 0.007--0.033).
Replacing RefSeq summaries with UniProt descriptions raised AUROC by 0.02--0.03 in every stratum with
linked publications (Supplementary Table~\ref{tab:uniprot} and Fig.~\ref{fig:uniprot}a,b).

For genes with 1--10 publications, UniProt descriptions named only 3.0\% of benchmark terms verbatim,
compared with 1.7\% for RefSeq and 29.5\% for generated summaries. Yet UniProt curators assigned
24\% of manual human GO annotations and 29\% of experimental annotations in GOA
\citep{huntley2015goa}. We examined this shared source in an exploratory analysis specified before
computation. For each gene we separately ranked terms assigned by UniProt and by other curators
against the same unannotated terms. UniProt-assigned terms tended to be more specific and ranked
lower across inputs. Among RefSeq-summary genes, the gap was 0.071 with the protein model and 0.073
with the symbol alone. Adding the RefSeq summary changed the gap by only $-0.002$ relative to the
protein model. Replacing RefSeq prose with UniProt prose improved the relative rank of UniProt-assigned
terms by 0.038 (0.031--0.044). For generated-summary genes, the same relative improvement over the
symbol alone was 0.075 (0.059--0.090) with UniProt prose and 0.060 with the generated summary
(Supplementary Fig.~\ref{fig:uniprot}c). Little exact overlap can therefore coexist with a measurable
association between the source of a description and the source of its labels.

\subsection{Removing generated summaries changes selected text--protein comparisons}\label{sec:without}

We next compared a text representation that included the gene symbol, full name, NCBI summary and
UniProt function description \citep{uniprot2025} with a protein sequence representation,
GB.Protein-16B \citep{sun2024aidoprotein}. For genes with no linked publications, rich text scored
0.038 above protein (0.024--0.053) with all summaries present. After removal the difference was
$-0.010$ ($-0.024$ to $+0.004$). With the MLP, the respective differences were $+0.015$ and
$+0.004$. An exploratory comparison with ESM2-650M \citep{lin2023esm2} gave linear-probe
differences of $+0.029$ ($0.018$ to $0.041$) and $-0.018$ ($-0.034$ to $-0.004$).

The labels help explain why the overall removal effect was relatively small in this stratum.
Under standard labels, 568 genes with no linked publications were evaluable and the removal contrast
was 0.027. After excluding phylogenetically inferred IBA annotations and reapplying the term-size
threshold, 100 genes were evaluable and the contrast was 0.088 (0.031--0.145). Of these, 90 were
also evaluable under standard labels. On those 90 the standard-label contrast was 0.091
(Supplementary Table~\ref{tab:zerobin}).

Many genes in the standard-label set have only broad, inferred terms that score well even with a
gene-independent frequency predictor.

Across all genes, filtered GenePT text scored slightly below GB.Protein-16B per gene
($-0.008$ linear, $-0.006$ MLP) and above it per term ($+0.048$, $+0.019$, post hoc).
Rich text retained a per-gene advantage ($+0.050$, $+0.024$) when UniProt descriptions
remained among its inputs (Supplementary Fig.~\ref{fig:supp}b). Throughout the paper,
\emph{summary-filtered} means that the Alliance-generated NCBI summaries were removed.
The remaining text can still draw on sources that contribute to GO.

\begin{table}[t]
  \centering
  \caption{\textbf{Per-gene GO AUROC by literature exposure.} Linear probe, out-of-fold; MLP probe in brackets for all
  genes. Columns are numbers of linked publications; 568, 4,310, 8,130, 3,279 and 414 genes with at least one benchmark
  term (16,701 in all); a gene's unannotated terms are its negatives. Rows marked $\dagger$ were added after
  the analysis plan; MLP values per bin in Supplementary Table~\ref{tab:mlp}. The term-frequency baseline scores every
  gene of a held-out fold with each term's frequency among the training-fold genes (no input from the gene).
  Within a fold its per-term AUROC is 0.5 when both classes are present; pooled across folds it is 0.433 because
  the training frequencies differ. These are frozen reported probe scores; GO tie handling awaits rescoring (Methods).}
  \label{tab:exposure}
  \small\setlength{\tabcolsep}{4pt}
  \begin{tabular}{@{}lcccccl@{}}
    \toprule
    Input & 0 & 1--10 & 11--100 & 101--1000 & $>$1000 & all \\
    \midrule
    GenePT text, as found                         & 0.962 & 0.897 & 0.882 & 0.862 & 0.837 & 0.883 [0.930] \\
    GenePT text, generated summaries removed      & 0.935 & 0.819 & 0.855 & 0.859 & 0.843 & 0.849 [0.920] \\
    GenePT text, replaced by LLM-written text$^\dagger$ & 0.929 & 0.812 & 0.852 & 0.859 & 0.840 & 0.846 [0.909] \\
    Gene symbol only                              & 0.936 & 0.843 & 0.844 & 0.848 & 0.849 & 0.848 \\
    Rich text, as found                           & 0.989 & 0.954 & 0.923 & 0.893 & 0.867 & 0.926 [0.956] \\
    Rich text, generated summaries removed        & 0.941 & 0.912 & 0.913 & 0.890 & 0.865 & 0.908 [0.950] \\
    Rich text, removed and gene names masked      & 0.814 & 0.865 & 0.907 & 0.886 & 0.864 & 0.888 [0.943] \\
    GB.Protein-16B                                & 0.951 & 0.865 & 0.859 & 0.835 & 0.796 & 0.857 [0.926] \\
    ESM2-650M$^\dagger$                            & 0.960 & 0.869 & 0.860 & 0.838 & 0.805 & 0.860 [0.929] \\
    Term-frequency baseline, no gene input$^\dagger$ & 0.861 & 0.808 & 0.793 & 0.783 & 0.768 & 0.796 \\
    \bottomrule
  \end{tabular}
\end{table}

\subsection{The effect persists across evidence filters and encoders}\label{sec:filter}

Restricting evaluation to experimental annotations retained the very sentences that most often
named the target. The Alliance gives experimental evidence priority when choosing terms
\citep{kishore2020alliance}. Under the experimental-evidence filter used by Zhong et al.\
\citep{zhong2025benchmark}, the removal contrast for genes with 1--10 publications was 0.098
(0.089--0.107) with the linear probe and 0.039 with the MLP. For experimental Biological Process
terms alone, it reached 0.118 (Fig.~\ref{fig:leak}c). The filter also changed which genes could
be evaluated. Only 41 genes without linked publications were evaluable with experimental labels, compared with 568 under standard labels.

Four additional neural encoders yielded contrasts of 0.088--0.092 in the 1--10 stratum. Even
TF-IDF yielded 0.043 (Fig.~\ref{fig:leak}d), so the input text alone can produce a difference
without a pretrained encoder. Across 4,949 GO terms, the per-term contrast increased with the
fraction of a term's genes whose generated summary named it (Spearman 0.23, permutation
$p<10^{-4}$, Fig.~\ref{fig:mechtime}a). Under experimental labels, 92\% of the contrast came
from sentences beginning with phrases such as “Enables”, “Involved in” or “Located in”.
Sentences beginning “Predicted to” contributed none. For genes without publications under the
standard labels, the latter sentences accounted for 94\%.

We had expected little change on endpoints derived independently of GO. Instead, generated
summaries improved MLP performance by about 0.02 on DepMap essentiality, gnomAD constraint and
GTEx tissue specificity (Fig.~\ref{fig:mechtime}b). The anticipated effect on HPA localisation
through HPA-derived GO annotations was small (0.007 with the linear probe). A description of a
gene's function can be informative on a separate assay even when its evaluation on GO is circular.

\subsection{Released embeddings and published benchmarks}\label{sec:published}

\begin{figure}[tp]
  \centering
  \includegraphics[width=\textwidth]{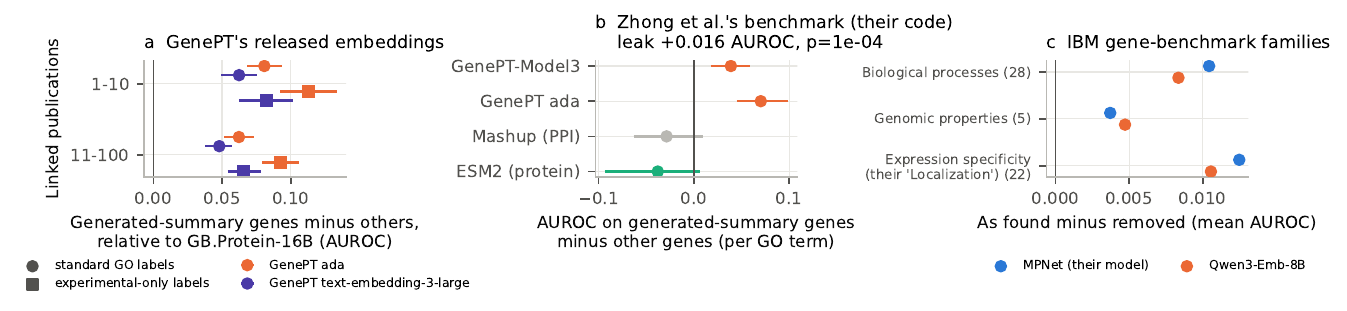}
  \caption{\textbf{Published embeddings and benchmarks.} (a) GenePT's released embeddings: generated-summary genes
  minus other genes, relative to GB.Protein-16B, under standard (circles) and experimental-only (squares) GO labels.
  (b) Zhong et al.'s benchmark rerun with their code: AUROC on generated-summary genes minus other genes per GO term,
  for four of their embeddings. (c) IBM gene-benchmark: mean AUROC gain from the generated summaries per task family
  (number of tasks in brackets) for their MPNet model and Qwen3-Embedding-8B.}
  \label{fig:published}
\end{figure}

The released GenePT embeddings provide an observational comparison. Within each exposure stratum
we measured the difference between genes with generated summaries and other genes, then subtracted
the same difference for GB.Protein-16B. The protein model had no access to the summaries. For
GenePT ada-002, the resulting contrast was 0.081 at 1--10 publications and 0.062 at 11--100.
Under experimental labels it was 0.113 and 0.092 (Fig.~\ref{fig:published}a). The corresponding
standard-label contrasts for text-embedding-3-large, which also encoded UniProt text, were 0.062
and 0.048. Genes that receive generated summaries may differ in other respects even within the
same exposure stratum, so these estimates are descriptive.

We also changed the text directly in two published evaluation protocols, retaining the authors'
code, splits and classifiers. In the 38-embedding benchmark of Zhong et al.\
\citep{zhong2025benchmark}, including generated summaries in GenePT-style text raised mean held-out
AUROC by 0.016 across 56 GO terms (Wilcoxon $p=0.0001$). Their released GenePT-Model3 scored
0.039 higher on genes with generated summaries than on other genes, while the network embedding
Mashup scored 0.029 lower on that group (Fig.~\ref{fig:published}b). The OMIM disease-gene task
changed little ($+0.0015$).

In the IBM gene-benchmark \citep{kantor2024genebenchmark}, we used the benchmark's MPNet encoder.
Including generated summaries raised mean AUROC by 0.010 across 28 biological-process tasks
(95\% bootstrap interval 0.007--0.014, 25 of 28 positive) and by 0.012 across 22 expression-specificity
tasks (0.011--0.014, all 22 positive). HPA biological-process annotations showed the largest gain
($+0.032$). The difference across five genomic-property tasks was 0.004 and was not statistically
significant (Fig.~\ref{fig:published}c). On GenePT's gene--gene interaction benchmark, where shared
GO annotations define the pairs \citep{du2019gene2vec}, removing the summaries changed AUROC by 0.005.

\subsection{Temporal evaluation must distinguish future text from prior annotations}\label{sec:temporal}

A temporal split raises two questions. Could a description contain an annotation added after
the cutoff? If the description predates the cutoff, does the task allow a model to read the gene's
earlier annotations? We examined these questions separately.

\begin{figure}[tp]
  \centering
  \includegraphics[width=\textwidth]{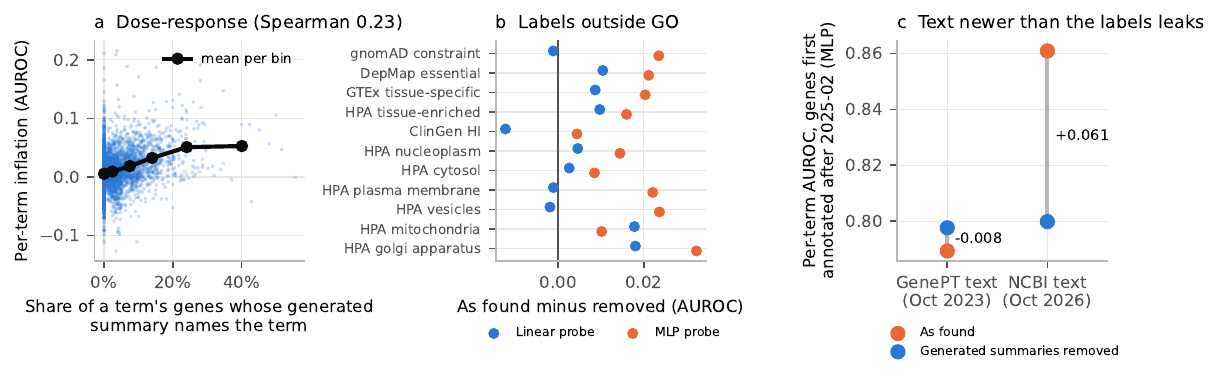}
  \caption{\textbf{Which annotations appear in the text, and when.} (a) Per-term inflation against the share of the term's genes whose generated
  summary names it (dots: GO terms, noisy individually; line: mean per bin, monotone). (b) Inflation on labels outside GO, linear and MLP probes.
  (c) Genes without a selected benchmark annotation in February 2025: GenePT's 2023 text and current NCBI text, as found and with generated
  summaries removed (MLP probe).}
  \label{fig:mechtime}
\end{figure}

For the first, we trained probes on February 2025 labels and evaluated genes that subsequently
acquired their first term \emph{within the benchmark vocabulary}. These genes could already have
annotations outside that vocabulary. With October 2026 NCBI text, per-term AUROC was 0.861 with
generated summaries and 0.800 without them. By then, the Alliance had incorporated later annotations
into the text. With GenePT's 2023 text, the corresponding values were 0.789 and 0.798
(Fig.~\ref{fig:mechtime}c). Protein models scored 0.855--0.868 on the same genes. Training-text
dates also mattered for protein models aligned with text. ProTrek \citep{su2024protrek} improved
on ESM2-650M \citep{lin2023esm2} by 0.055 for annotations already present when its training text
was written, and by 0.005 for genes newly represented in the benchmark. ProtST \citep{xu2023protst}
retained a 0.023 advantage over ESM1b \citep{rives2021esm1b} on the latter genes (Supplementary
Fig.~\ref{fig:supp}a).

For the second question, we used the temporal task of Zhong et al. It scores experimental
annotations added after March 2024 for genes that may already have annotations. Our rerun
recovered their published mean scores for GenePT-Model3 (0.774) and Mashup (0.776).
Removing generated summaries lowered AUROC by 0.027 with 2026 text and by 0.024 with 2023
text ($p=0.001$ for the latter), even though none of the positive genes' 2023 generated
summaries named its held-out term. The larger change came from the choice of negative genes.
Against negatives with a generated summary, AUROC fell from 0.744 to 0.660 on removal.
Against other negatives it fell from 0.662 to 0.651. An older summary can help a classifier
rule out genes by revealing what was already known about them. For tasks that permit this
information, a natural comparison is a model given the gene's annotation vector at the cutoff
(Section~\ref{sec:know}).

\subsection{Descriptions written by two language models show little template overlap}\label{sec:llm}

scELMo released two sets of embeddings made with the same encoder. One used descriptions
from NCBI, the other descriptions written by GPT-3.5 \citep{liu2023scelmo}. We applied the
observational contrast from Section~\ref{sec:published} to each. For the NCBI text the contrast
was 0.057 (0.045--0.069) at 1--10 publications and 0.042 (0.033--0.051) at 11--100.
For the GPT-3.5 descriptions it was $-0.034$ ($-0.048$ to $-0.021$) and $-0.016$
(Fig.~\ref{fig:beyond}b). With the same prompt, Qwen2.5-7B-Instruct \citep{qwen2024qwen25}
descriptions encoded by Qwen3-Embedding-8B gave $-0.038$ and $-0.020$. Among generated-summary
genes with 1--10 publications, Qwen2.5 descriptions named 1.5\% of the gene's benchmark GO
terms verbatim, compared with 31.4\% for the generated summaries. None of the 19,256 Qwen2.5
descriptions contained an Alliance template sentence. Excluding the 40 descriptions partly in
Chinese changed no contrast by more than 0.001 (post hoc; Methods).
These two models showed little evidence of copying the templates under the prompt tested.
Other forms of overlap, including memorised annotations, remain possible.

Among genes with 1--10 publications, GB.Protein-16B scored above the GPT-3.5 and Qwen2.5
descriptions by 0.037 (0.030--0.044) and 0.062 (0.055--0.069), respectively
(Fig.~\ref{fig:beyond}c, post hoc). ESM2-650M gave differences of 0.041 and 0.065.
GPT-3.5 text overtook GB.Protein-16B above 100 linked publications, and Qwen2.5 text above
1,000. In the temporal test on genes with at most five publications, GB.Protein-16B's advantage
over frozen GPT-3.5 descriptions was 0.017 ($-0.003$ to $0.037$) with the MLP, below the
prespecified threshold.

\subsection{Generated summaries in six model organisms}\label{sec:organisms}

Alliance descriptions are especially common outside human \citep{ncbi2022insights}. Among NCBI
protein-coding genes, they cover 70\% in mouse, 76\% in rat, 62\% in zebrafish, 86\% in fly,
64\% in worm and 95\% in yeast. Among genes with 1--10 linked publications, coverage ranged
from 90\% to 98\% across these species. The corresponding human estimates were 30\% overall
and 57\% at 1--10 publications. These counts use NCBI's gene list for all species. The 32.1\%
reported in Section~\ref{sec:prev} uses the HGNC human set (Supplementary Table~\ref{tab:organisms}).
RefSeq summaries covered 10\% of mouse genes, 6\% of rat genes and less than 1\% in each
of the other four species.

\begin{figure}[tp]
  \centering
  \includegraphics[width=\textwidth]{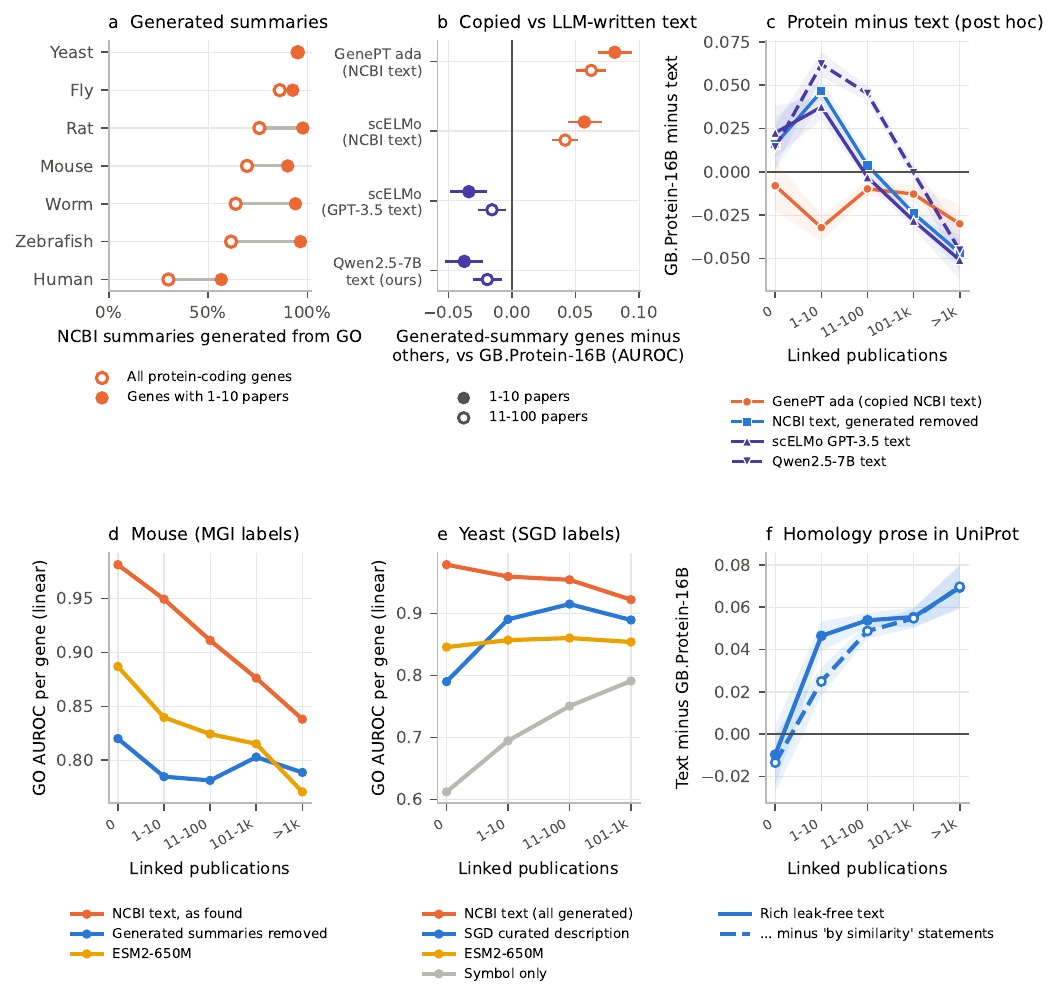}
  \caption{\textbf{Summary source, literature exposure and organism.} (a) Share of protein-coding genes whose NCBI Gene summary is
  generated from GO, over all genes (open) and for genes with 1--10 publications (filled). (b) Generated-summary genes
  minus other genes, relative to GB.Protein-16B, for two published embeddings of NCBI summaries (GenePT and scELMo,
  both ada-002) and two sets of LLM-written descriptions (scELMo's GPT-3.5 descriptions; ours from Qwen2.5-7B-Instruct),
  for genes with 1--10 (filled) and 11--100 (open) publications; 95\% bootstrap intervals. (c) Post hoc: GB.Protein-16B
  minus each text embedding by literature exposure (per-gene GO AUROC, linear probe; bands are 95\% intervals).
  (d) Mouse, MGI labels: NCBI text as found and with generated summaries removed, and ESM2-650M. (e) Yeast, SGD labels:
  NCBI text (all generated), SGD's curator-written descriptions, the symbol alone, and ESM2-650M. (f) Rich summary-filtered
  text minus GB.Protein-16B, with and without UniProt function statements supported only by sequence-based inference.}
  \label{fig:beyond}
\end{figure}

We tested predictions in two of the six organisms. For mouse genes with 1--10 publications,
removing generated summaries lowered per-gene AUROC against MGI annotations by 0.165
(0.160--0.170) with a linear probe and 0.080 with the MLP (Fig.~\ref{fig:beyond}d).
Across all evaluated mouse genes, the original text scored 0.094 above ESM2-650M.
Removal left 91\% of genes with a symbol alone and put text 0.045 below ESM2-650M overall.

In yeast, all available NCBI summaries were generated. NCBI text reached 0.954, compared
with 0.902 for descriptions written by Saccharomyces Genome Database (SGD) curators, 0.858
for ESM2-650M and 0.731 for symbols alone (Fig.~\ref{fig:beyond}e). The NCBI--SGD difference
was 0.052 (0.048--0.055) overall and 0.069 for genes with 1--10 publications. Among the 81
open reading frames without linked publications, ESM2-650M scored 0.846 and SGD text 0.790.
That difference was not statistically significant. SGD curates both the prose and the yeast
GO labels, a relationship similar to the one we found for UniProt and human GOA.

\section{GeneProv compares gene representations across tasks}\label{sec:bench}

\subsection{Tasks, tracks and scoring}\label{sec:benchdesign}

GeneProv asks how a representation performs when the source of the target changes
(Table~\ref{tab:design}). Its four tasks cover GO annotation recovery (T1), eleven non-GO
endpoints evaluated with gene families held out (T2), experimental annotations added after
a January 2023 cutoff (T3), and retrieval of co-essential partners by embedding similarity (T4).
We report each task for the full gene set and for genes with few publications. The primary
stratum is 0--10 linked publications, except on T3, where we count 0--5 at the cutoff.

Track F fixes the text for every encoder. Each encoder receives the GenePT corpus twice, with
and without generated summaries. Its score uses the filtered text, and we report the paired
removal contrast separately. Track O compares representations as supplied. It includes both
sequence models and released embeddings that retain generated summaries. A Track O comparison
therefore holds each model's inputs fixed as the evaluation task changes.

For the GeneProv Index, we transform each task AUROC to $100\,(2\,\mathrm{AUROC}-1)$ and
average T1, T2 and T4 with equal weight. A score of 0 corresponds to random ranking, 100 to
perfect discrimination, and $-100$ to a fully reversed ranking. T3 is reported separately
because only some representations satisfy its annotation-access rule. A given model's score
can be calculated independently of the other entries. Its rank depends on which models are included.

Three properties matter when interpreting the index. First, the random-ranking reference is
well below the term-frequency baseline on T1. Second, the component tasks cover different sets
of genes, and the T2 class-size rule can change which labels enter the average across strata.
Third, T2 essentiality and T4 co-essentiality both use DepMap measurements. The component
scores show which tasks account for a difference between representations.

\begin{table}[tp]
  \centering
  \caption{\textbf{GeneProv v1 at a glance.} The final column identifies the comparison that motivated each design choice.}
  \label{tab:design}
  \footnotesize\setlength{\tabcolsep}{4pt}
  \begin{tabular}{@{}p{0.15\textwidth}p{0.52\textwidth}p{0.29\textwidth}@{}}
    \toprule
    Component & Definition & Motivation \\
    \midrule
    Universe, strata & 19,256 HGNC protein-coding genes; five folds shared by every row; headline stratum genes
      with 0--10 publications (7,120), all genes beside it & exposure dependence (Section~\ref{sec:inflate}) \\
    T1, GO recovery & 4,949 GO terms (no IEA or ND); Track F removes generated summaries; Track O uses declared inputs; per-gene AUROC,
      linear and MLP probes & generated summaries (Sections~\ref{sec:prev}--\ref{sec:filter}); still coupled to
      UniProt curation (Section~\ref{sec:prov}) \\
    T2, non-GO labels & eleven assay, population and curated endpoints (DepMap essentiality, gnomAD constraint, ClinGen dosage,
      six HPA locations, GTEx and HPA tissue specificity; the essentiality label and T4 share the DepMap source;
      ClinGen drops out of the 0--10 stratum by the class-size rule); folds grouped by HGNC family & labels outside GO
      (Section~\ref{sec:filter}); homology between family members \\
    T3, annotation holdout & experimental annotations added after 1 January 2023 to genes without experimental annotation
      then (such genes may carry inferred annotations; 137 genes with 0--5 publications at the cutoff); only
      representations eligible under the annotation-access rule, which admits later sequence/expression models (Methods)
      & temporal holdouts (Section~\ref{sec:temporal}) \\
    T4, co-essentiality & retrieval of each gene's ten DepMap co-essential partners by embedding similarity; no probe
      & DepMap-derived endpoint \\
    GeneProv Index & mean over T1, T2 and T4 of $100\,(2\,\mathrm{AUROC}-1)$; T3 beside it; paired bootstrap
      intervals for every index and rank; transform independent of the other rows & declared task aggregation; task-level scores retained \\
    Track O & any representation; declares the provenance of embedded text (share generated from GO by tag or
      detector, text used in training, mixture); twenty rows in v1 & provenance tags and the detector
      (Sections~\ref{sec:prev}, \ref{sec:published}) \\
    Track F & the benchmark supplies GenePT's 2023 text with and without generated summaries; the embedder is scored
      on the summary-filtered corpus and reports $S$; six embedders in v1 & encoders (Section~\ref{sec:filter}), released
      embeddings (Section~\ref{sec:published}) \\
    Controls & symbol-only, name-masked and LLM-written rows in the table & name prior (Section~\ref{sec:without}),
      LLM-written text (Section~\ref{sec:llm}) \\
    \bottomrule
  \end{tabular}
\end{table}

\input{table_bench_v1}

\begin{figure}[tp]
  \centering
  \includegraphics[width=0.92\textwidth]{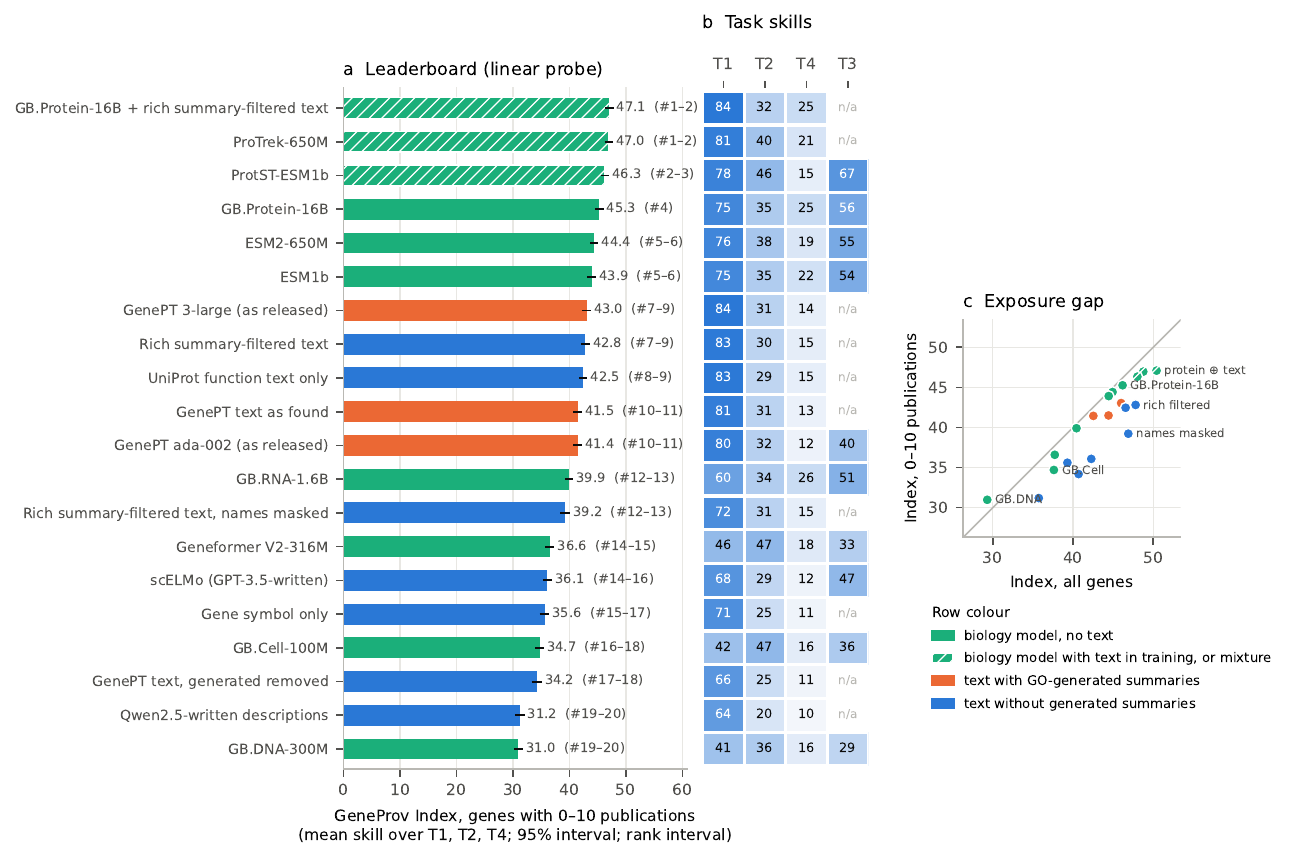}
  \caption{\textbf{GeneProv v1 leaderboard.} (a) Headline GeneProv Index (genes with 0--10 publications, linear
  probe) with 95\% bootstrap intervals and rank intervals; bars are coloured by text provenance, hatched rows embed no
  text but used UniProt function text in training, or are a declared mixture. (b) The per-task skills behind each
  index ($100\,(2\,\mathrm{AUROC}-1)$; T3 for representations eligible under the 2023 annotation-access rule, genes with 0--5
  publications then). (c) The same index on all genes against the headline; points above the diagonal have higher
  scores on the all-gene population.}
  \label{fig:leaderboard}
\end{figure}

\subsection{Rankings depend on task composition and literature exposure}\label{sec:leaderboard}

For genes with 0--10 publications, all three protein sequence models scored above the best text-only
model on the original linear-probe index. GB.Protein-16B scored 45.3 (95\% gene-bootstrap interval
44.5--46.0), ESM2-650M 44.4 and ESM1b 43.9. GenePT text-embedding-3-large scored 43.0
(Fig.~\ref{fig:leaderboard}, Table~\ref{tab:bench}). The highest scores belonged to a concatenation
of protein and rich text embeddings (47.1) and two protein models trained with text, ProTrek
(47.0) and ProtST (46.3). With all genes included, rich summary-filtered text scored 47.8 and
GB.Protein-16B scored 46.2. Like their T1 component, these indices await tie-aware rescoring.

Changing the task mix changed which representations ranked highest. With GO omitted, the
single-cell representations led and the rank correlation with the three-task index was only
0.43. Each representation used exactly the same inputs for both rankings
(Supplementary Fig.~\ref{fig:taskranks}). The metric within a task mattered too. On co-essentiality,
the ranking by AUROC had a correlation of only 0.03 with the ranking by recall at ten. A model
that ranked true partners above most genes could still place few of them among its first ten.
Removing UniProt from the protein--text concatenation lowered its index from 47.1 to 42.9.
This change removed both functional information and text from a source that contributes GO labels.

Resampling gene families together widened the interval for GB.Protein-16B from 44.5--46.0 to
40.1--49.9. The primary 1--10-publication removal contrast retained a positive interval
(0.070--0.086). Supplementary Tables~\ref{tab:cluster} and \ref{tab:coverage} report the
sensitivity to related genes and incomplete coverage.

\subsection{What prior annotations explain in the temporal task}\label{sec:know}

\begin{figure}[tp]
  \centering
  \includegraphics[width=0.74\textwidth]{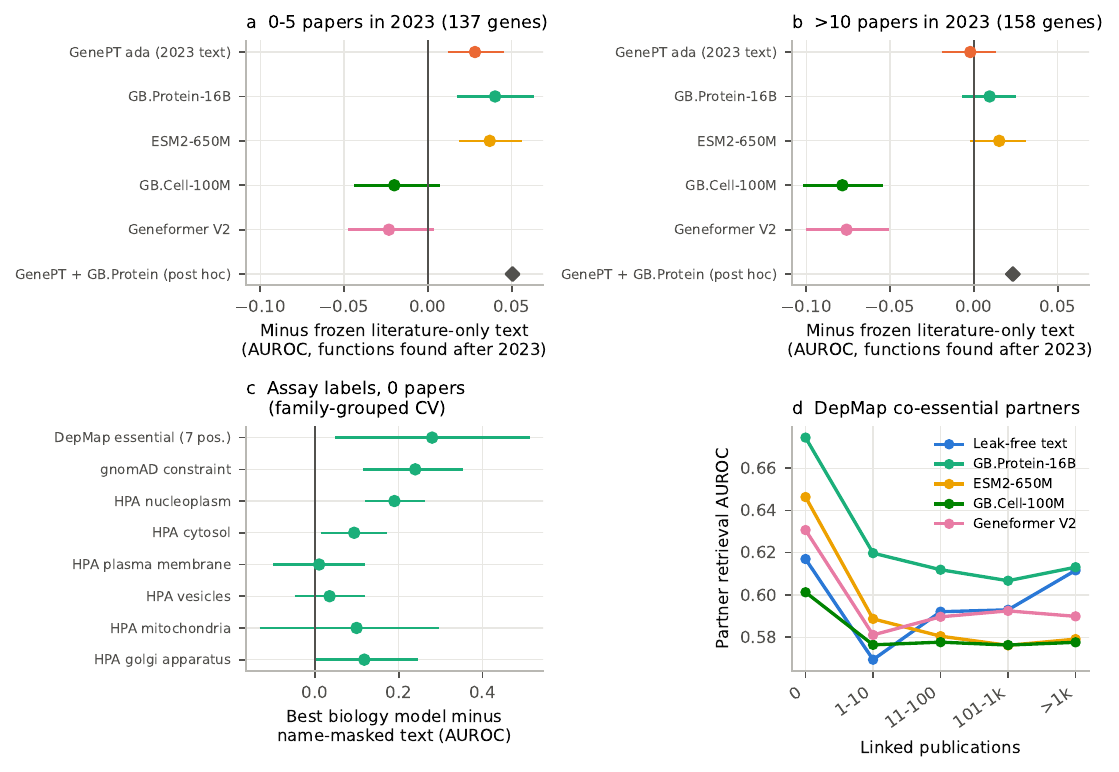}
  \caption{\textbf{Prediction for genes with few linked publications.} (a,b) Experimental annotations added after 1 January 2023 for
  genes without experimental annotation then: each representation minus literature-only text frozen at the cutoff, for
  genes with 0--5 (a) and more than 10 (b) publications at the cutoff (name-masked frozen text: $-0.240$ and $-0.119$);
  bars are 95\% bootstrap intervals; diamonds mark the post hoc average of GenePT and GB.Protein-16B scores.
  (c) Genome-wide assay labels for genes without publications under family-grouped cross-validation: best biology model
  minus name-masked text. (d) Retrieval of DepMap co-essential partners by embedding similarity.}
  \label{fig:silent}
\end{figure}

T3 evaluates genes that lacked experimental annotations in January 2023 and subsequently gained
them. Among the 137 evaluation genes with at most five publications at the cutoff, MLP AUROC was
0.917 for GB.Protein-16B, 0.914 for ESM2-650M and 0.877 for the frozen literature-only text
(Fig.~\ref{fig:silent}a). We then excluded target terms already assigned to another member of
the same HGNC family at the cutoff. That left 627 of 1,172 targets across 99 genes in this stratum,
and 1,261 of 3,224 targets across all 347 evaluation genes. The family-restricted advantage of
GB.Protein-16B over frozen text was 0.067, 0.029--0.106 by paired bootstrap. Expression models
showed no comparable advantage on the primary targets, and the protein advantage over frozen text
disappeared among genes with more than ten publications (Fig.~\ref{fig:silent}b).

GenePT's 2023 embeddings were a stronger comparator. They included Alliance sentences derived
from sequence-based predictions for many of these genes. GB.Protein-16B's advantage was 0.012
and was not statistically significant. On the family-restricted targets it was 0.020
($-0.017$ to $0.057$). The planned test against GenePT therefore failed. The planned combination
of standardised model scores also failed its threshold. In a post hoc comparison, averaging the
two raw scores gave 0.928, 0.022 above GenePT. For genes without publications, the best biological
representation scored above name-masked text on five of eight assay labels (four with GB.Protein-16B
fixed). It also scored higher on co-essentiality retrieval (0.674 versus 0.617, Fig.~\ref{fig:silent}c,d).

Prior annotations provide a direct reference point. Of the 347 evaluation genes, 260 already had
inferred annotations, and 35\% of the 3,224 future experimental targets had inferred support in
2023. For genes with 0--5 publications, the binary annotation vector scored 0.493 with a linear
probe and 0.889 with an MLP. The MLP score exceeded frozen text (0.877) and fell 0.028 below
GB.Protein-16B (0.005--0.055). On targets without prior inferred support the latter difference was
0.035 (0.000--0.072). Adding the vector to GB.Protein-16B yielded no further gain
($-0.020$, $-0.040$ to $0.000$, post hoc, Supplementary Table~\ref{tab:v11}).

Sequence-based inferences also appear in UniProt prose. For 12\% of genes without publications,
21\% of those with 1--10 and 5\% of those with 101--1000, every UniProt function statement rested
solely on sequence evidence. Removing such statements from rich summary-filtered text lowered
per-gene AUROC by 0.022 (0.018--0.025) at 1--10 publications and by 0.100 among the genes that lost
a statement. The text advantage over GB.Protein-16B in this stratum fell from 0.047 to 0.025
(Fig.~\ref{fig:beyond}f). With experimental labels, the removal contrast was only 0.003.

\subsection{Using GeneProv}\label{sec:rules}

For a new text encoder, Track F asks how well it performs on the filtered corpus and how much
its score changes when the generated summaries are restored. For a released embedding or a
model trained on multiple data sources, Track O records which genes it covers and where its
inputs and training text came from. Missing genes receive the mean embedding. The benchmark
website (\url{https://hwang595.github.io/gene-embedding-exposure/}) presents the task profiles,
uncertainty intervals and provenance declarations together. The v1 record fixes the original
comparisons. Supplementary Table~\ref{tab:v11} identifies the controls appended in v1.1.

\section{Discussion}

For a gene with few publications, the most informative sentences in its NCBI summary may have
been written by an annotation pipeline. This matters most when the task is to recover the
annotations that pipeline used. In GenePT's released text, the source tags have disappeared
but the sentences remain. Removing them produced the largest reported change among genes with
1--10 linked publications. The audit establishes where the sentences came from independently
of the GO scoring error. The size of the performance change remains provisional.

Earlier work described circularity in function prediction and the risk of recovering GO terms
from generated descriptions \citep{gillis2013,joachimiak2023spindoctor}. Here we could trace
that risk to individual model inputs and vary those inputs in the same evaluation. Even TF-IDF
registered a change when the summaries were removed. Restricting labels to experimental
evidence preserved many of the terms the Alliance had chosen to express in sentences. At the
same time, the summaries helped predict several non-GO endpoints. Retaining a description
can be appropriate when the intended application has access to it and the target comes from
an independent measurement.

UniProt illustrates why searching for exact term names is insufficient. Its descriptions
rarely reproduced the wording of the GO term, yet they recovered most of the performance lost
on removal among genes for which a replacement was available. The terms that benefited most,
relative to other terms on the same gene, were those assigned by UniProt curators. Shared
curation can transmit relevant knowledge even when a sentence was written independently.
The replacement experiment cannot partition the gain between shared evidence and other
biological detail in the text. The LLM-written descriptions showed little evidence of the
Alliance template, although memorisation remains possible for both the generator and the encoder.

The relevant cutoff in a temporal benchmark depends on the question. Someone completing an
existing annotation record may reasonably consult its earlier entries. Someone predicting
the first experimental evidence for a gene may have only inferred annotations to work with.
A comparison should say which of these inputs each model receives. Our T3 task dates the
appearance of experimental annotations in GO. The experiments themselves may have taken place
earlier, and the protocol admits sequence and expression models released after the cutoff.
Its scores describe a retrospective annotation holdout under those access rules.

Several features of the evaluation limit what can be inferred about understudied genes.
A count of zero means that no qualifying PubMed article is linked to the gene. Database links
can be incomplete, and the historical exposure count uses an identifier threshold to approximate
publication date. Unannotated pairs serve as negatives even when the function is unknown.
In the no-publication stratum, excluding phylogenetically inferred IBA annotations changed the
number of evaluable genes from 568 to 100. Related genes also share information. Grouping
them in the bootstrap substantially widened some index intervals. Only 137 genes contributed
to the primary T3 stratum, and the zero-publication DepMap analysis included just seven essential
genes. Outside human, we tested predictions in mouse and yeast. Our experiments do not address
downstream cell or perturbation-response tasks.

The removal contrast includes changes in probe training as well as evaluation. Text replacements
differed in length and factual coverage. The gene bootstrap conditions on a fitted probe, and
GO propagation through regulatory relationships can group a regulator with a process in which
it does not directly participate. A later attempt to regenerate the benchmark exceeded the
planned tolerance in 16 of 366 cells. Seven ambiguous GenePT aliases also require a sensitivity
analysis. Most importantly, the historical GO scorer treated equal scores as ordered. Without
the original predictions, the effect on GO contrasts and composite ranks is unknown (Methods).

GeneProv offers two comparisons with different interpretations. Changing the corpus for a fixed
encoder measures the contribution of generated summaries to the evaluation. Changing the tasks
for a fixed representation shows what its score rewards. A reader deciding whether to use an
embedding to prioritise genes for an assay needs the assay-specific result, the model's coverage
of genes like those of interest, and an account of what the model could already know about them.

\section{Methods}

\subsection{Genes, descriptions and encoders}

\paragraph{Genes and literature exposure.}
We studied 19,256 HUGO Gene Nomenclature Committee (HGNC) protein-coding genes with Ensembl identifiers \citep{seal2023hgnc}.
For each gene we counted distinct articles linked by NCBI gene2pubmed, retaining articles linked
to at most 20 human genes. We grouped genes into five exposure strata (0, 1--10, 11--100,
101--1000 and $>$1000 articles). To approximate exposure on 1 January 2023 we counted PubMed
identifiers below 36,594,000. For comparison, the median identifier created on 3 January 2023
was 36,595,358. This proxy dates database records approximately. A count of zero means that no
qualifying link was found.

\paragraph{Summary provenance.}
We retrieved NCBI Gene summaries with E-utilities \texttt{esummary} on 6 October 2026
\citep{sayers2025ncbi} and took the source from each summary's last bracketed tag. GenePT's
October 2023 archive retains those tags on full NCBI Gene pages \citep{genept2023zenodo}.
Its March 2024 NCBI-only text omits the tags \citep{genept2024zenodo}. We assigned each 2024
entry the provenance of the same key in 2023. We matched study genes to GenePT keys by current
symbol, then alias and previous symbol. Seven ambiguous matches are discussed below.
To identify a generated summary without its tag, we checked whether the first sentence began with
an Alliance template phrase. We added “Predicted to be a/an” after inspecting the results.

\paragraph{Text inputs.}
Our basic input was the GenePT string “Gene Symbol X” followed by the NCBI summary. We used
either the 2024 GenePT release or the 2026 NCBI snapshot. Filtering removed only the summary
of an Alliance-tagged gene. In the matched-removal control we sampled RefSeq-summary genes
with replacement (seed 0), matching each generated-summary gene with 1--100 publications by
exposure stratum and GO-term-count decile. For the LLM replacement we substituted its
description embedding for the original row. Each gene was encoded independently, so this
gives the embedding of the hybrid corpus. Within Alliance summaries, we also removed either
the predicted sentences or the asserted GO sentences. Rich inputs contained symbol, full name,
NCBI summary and UniProt function text. Name masking removed symbols, aliases and previous
symbols from the filtered rich text.

For the 2023 literature-only reference we took GenePT's October 2023 text, removed generated
summaries, and also removed nine summaries dated January 2023 or later. Only the tag date was
available to approximate the text at the cutoff. Symbols and aliases came from the 2023 keys.
We used all-mpnet-base-v2, trained before the cutoff, for both the masked and unmasked versions.

\paragraph{Representations.}
We used Qwen3-Embedding-8B with bf16 weights, a 2,048-token limit and L2 normalisation
\citep{zhang2025qwen3embedding}. The other neural text encoders were Qwen3-Embedding-0.6B,
all-mpnet-base-v2 \citep{song2020mpnet,reimers2019sbert}, bge-large-en-v1.5 \citep{xiao2023cpack}
and pubmedbert-base-embeddings \citep{gu2021pubmedbert}. For TF-IDF we used word unigrams and
bigrams, minimum document frequency two, sublinear term frequency and a 1,024-component truncated
SVD \citep{pedregosa2011sklearn}. We fitted its vocabulary and SVD once on the two corpora pooled
without labels. We also evaluated GenePT's released ada-002 and text-embedding-3-large vectors
\citep{chen2024genept}. The latter include UniProt text.

We mean-pooled GB.Protein-16B over its 2,048-token context \citep{sun2024aidoprotein} and
mean-pooled ESM2-650M and ESM1b after cropping proteins to 1,022 residues
\citep{lin2023esm2,rives2021esm1b}. The human GO comparison with ESM2-650M was added after
planning. For GB.Cell-100M \citep{ho2024aidocell} we averaged gene-token states over the cells
expressing each gene in a sample of 21,000 Tabula Sapiens cells \citep{tabulasapiens2022}.
We used token embeddings for Geneformer V2-316M \citep{theodoris2023geneformer,chen2024geneformerv2}.
The text-aligned protein models were ProTrek-650M \citep{su2024protrek} and ProtST-ESM1b
\citep{xu2023protst}. Supplementary Table~\ref{tab:models} gives training sources and coverage.

\subsection{Targets and evaluation}

\paragraph{GO and non-GO labels.}
We used the 2026 human GOA annotations \citep{ashburner2000go,go2023,huntley2015goa}, excluded annotations with a
NOT qualifier and propagated terms with GOATOOLS \citep{klopfenstein2018goatools}. Specifically,
\texttt{get\_all\_upper()} traversed \texttt{is\_a} and the loaded relationships, including
\texttt{part\_of}, \texttt{regulates}, \texttt{positively\_regulates} and
\texttt{negatively\_regulates}. This broad rule can group a regulator with a process in which
it does not directly participate \citep{gorelations}. We retained terms assigned to 20--1000
genes. Excluding IEA and ND evidence gave the standard set of 4,949 terms. Variants included
IEA, excluded IBA, kept only experimental evidence (EXP, IDA, IPI, IMP, IGI, IEP, HTP, HDA,
HMP, HGI and HEP, 3,531 terms), or restricted the experimental set to Biological Process
(2,569 terms).

The eleven non-GO endpoints were ClinGen haploinsufficiency \citep{rehm2015clingen}, gnomAD v4.1
LOEUF below 0.6 \citep{karczewski2020gnomad}, DepMap 24Q4 common essential versus non-essential
genes \citep{tsherniak2017depmap,depmap24q4}, six HPA main subcellular locations \citep{thul2017hpa},
GTEx tissue specificity \citep{gtex2020} and HPA tissue enrichment \citep{uhlen2015hpa}.
For co-essentiality, each gene's partners were the ten other genes with the highest correlation
in CRISPR gene effect across 1,178 DepMap cell lines.

\paragraph{Temporal targets.}
For the February 2025 holdout we trained on GO labels from 6 February 2025 and scored annotations
added subsequently. We defined a gene's prior annotation status within the selected term vocabulary.
For T3 we trained on the 1 January 2023 release under the standard evidence rule, which admits
inferred annotations. We evaluated genes with no experimental annotation at that cutoff. A target
was a gene--term pair with experimental evidence by 2026 for a term present in both releases.
We dated the addition of evidence from GO releases. The novel-pair analysis further excluded any
target with a non-IEA annotation in 2023.

\paragraph{Probes and uncertainty.}
We assigned the mean vector to genes lacking an embedding, then standardised each dimension
across all genes. This step, the pooled TF-IDF/SVD fit and the full-universe term vocabulary
use information about held-out genes. Probe labels were held out by fold. For multilabel
prediction we fitted logistic regression with full-batch AdamW for 300 epochs (learning rate
0.01, weight decay $10^{-4}$), or a one-hidden-layer MLP with 512 GELU units, dropout 0.2 and
the same schedule. Binary linear probes used 500 epochs and weight decay $1/n_{\mathrm{train}}$.
Binary MLPs used the multilabel MLP schedule. We fixed all settings in the first analysis round.

Each representation used the same five shuffled folds (seed 0) within a task. Family-grouped
folds used the first HGNC group for genes with multiple memberships and treated ungrouped genes
as singletons. We calculated metrics from out-of-fold scores. For paired differences, we sampled
genes with replacement 1,000 times and report 95\% bootstrap intervals. These intervals condition
on the fitted probes and do not adjust for multiple comparisons. Supplementary Table~\ref{tab:cluster}
reports intervals with HGNC families resampled together. Section~\ref{sec:inflate} reports the
separate check that redrew the folds and refitted the probes.

\paragraph{Co-essentiality retrieval.}
We ranked candidate genes with DepMap partners by cosine similarity of the standardised embeddings.
We excluded the query itself, retained mean-imputed candidates and assigned average ranks to ties.
Supplementary Table~\ref{tab:coverage} reports recall at ten alongside AUROC. The mixture
concatenated the standardised protein and text blocks without reweighting. Its 4,096 text
dimensions outnumber its 2,304 protein dimensions by a factor of 1.8. Both blocks contribute
directly to cosine similarity, whereas a fitted probe can learn separate dimension weights.

\paragraph{Published benchmark reruns.}
For Zhong et al.\ \citep{zhong2025benchmark}, we used commit \texttt{d132002}, the released splits,
support-vector classifiers and $C$ grid, and the 11,327 genes shared by their intersection set and
our study. Changes were limited to file paths, saving per-gene held-out scores and calling
\texttt{main()} in the temporal script. We recovered the published mean AUROC for each method.
Individual term scores correlated only about 0.5 with the published values, for reasons we
could not establish. We therefore made all comparisons within the rerun. For the IBM gene-benchmark
\citep{kantor2024genebenchmark}, we used commit \texttt{d2b345e}, built tasks with its scripts,
fetched the Reactome files separately, seeded NumPy and applied the documented multilabel threshold
of 0.1. For gene--gene interactions \citep{du2019gene2vec} we used the classifier in GenePT's notebook.

\subsection{Additional controls}

\paragraph{LLM-written descriptions.}
We matched scELMo's released 1,536-dimensional embeddings of GPT-3.5 descriptions to 19,231 genes
by symbol, alias or previous symbol \citep{liu2023scelmo}. We prompted Qwen2.5-7B-Instruct
\citep{qwen2024qwen25} with the scELMo prompt, “Please summarize the major function of gene: X.
Use academic language in one paragraph and include pathway information.” We used the chat template,
bf16, greedy decoding and at most 256 new tokens. The median description was 140 words. We kept
all 40 partly Chinese responses (0.2\%) and encoded the output as written with Qwen3-Embedding-8B.
For verbatim overlap we measured the share of a gene's benchmark GO terms whose names occurred
in its lower-cased description.

\paragraph{UniProt substitutions and evidence.}
We replaced generated summaries for 4,591 genes and RefSeq summaries for 11,074 genes with the
corresponding UniProt function text, keeping the “Gene Symbol X” prefix. We evaluated the
substituted genes in each comparison. To identify sequence-derived statements, we took the prose
preceding each evidence block together with its terminating full stop. We removed statements
supported exclusively by ECO:0000250 (by similarity) or ECO:0000255 (sequence model). This affected
2,652 genes. We used the same encoder for both versions.

For the co-curation analysis, we split direct benchmark GOA annotations by the
\texttt{assigned\_by} field into UniProt and other contributors, after excluding IEA, ND and NOT.
For each of 8,338 genes with terms from both groups, we calculated two out-of-fold linear-probe
AUROCs against the same unannotated terms. We averaged the within-gene difference and paired
genes when comparing inputs.

\paragraph{Alternative probes, folds and targets.}
For the nearest-neighbour probe we used the 20 closest training genes by cosine similarity on
standardised embeddings and averaged their label vectors with softmax weights (temperature 0.1).
We redrew the five folds with seeds 1--3. For the family-restricted temporal analysis we retained
only target terms absent from every other gene in the same HGNC family in the 2023 training labels.
We kept the original negative set and used the saved out-of-fold scores.

\paragraph{Model organisms.}
We obtained protein-coding genes and summary sources from the October 2026 NCBI gene records.
We counted gene2pubmed articles linked to at most 20 genes of the same organism. The mouse
evaluation included 16,421 protein-coding genes with a reviewed UniProt entry (longest entry
per gene) and 5,691 MGI terms \citep{mgi2024}. In yeast, 5,929 SGD open reading frames had both
a reviewed UniProt entry and an NCBI Gene record. We used 1,631 SGD terms and the curator-written
SGD descriptions \citep{sgd2022}. Both organisms used the standard evidence, propagation and
term-size rules. We encoded the symbol and summary with Qwen3-Embedding-8B and the sequence with
ESM2-650M (mean-pooled, up to 1,022 residues). We fitted probes and calculated intervals as in human.

\subsection{Benchmark construction and validation}

\paragraph{Scoring and eligibility.}
T1 uses the standard GO labels. T2 averages the within-stratum AUROCs of endpoints with at least
five genes of each class, using family-grouped folds. This excludes ClinGen in the 0--10
stratum. T3 uses the primary experimental holdout and T4 uses the co-essential partners defined
above. The index averages T1, T2 and T4. For each of 1,000 paired bootstrap samples (seed 0),
we recalculated all task means, including the mean over eligible T2 labels, and ranked all
Track O entries. We also tabulated a four-task index for T3-eligible models. For Track F,
$S$ denotes the per-gene GO removal contrast on genes with 1--10 publications.

For T3 we checked whether a model's declared inputs or their date proxies indicated functional
annotation text newer than 1 January 2023. The protocol admitted sequence and expression models
regardless of training date, including GB.Protein-16B, released in 2024. For GenePT ada-002 we
used the pre-2023 summary tags on the evaluation genes, although the archive was collected in
October 2023. For scELMo we used GPT-3.5's reported 2021 training-data cutoff. ProtST's Swiss-Prot
2022 text qualified, while ProTrek's Swiss-Prot 2024 text did not. Frozen literature-only text
encoded with MPNet supplied the reference (Supplementary Table~\ref{tab:models}). These proxies
limit how closely T3 approximates a prediction made at the cutoff.

\paragraph{Benchmark versions and regeneration.}
Version 0 fixed the tasks, representations and exposure strata. Version 1 added the index, seven
Track O entries, and T2 and T4 for Track F. Every v0 cell reproduced within 0.0005 (218 of 218).
The three planned ranking checks also passed. Among unmixed models without text alignment, a
text-free model led the linear-probe index. Both GenePT releases placed at least two positions
lower on the index than on T1, and GB.DNA-300M ranked last among the text-free models. The first
implementation of the first ranking check included text-aligned models. We logged and corrected
the discrepancy before rerunning it (Supplementary Table~\ref{tab:rounds}).

After loss of the original scratch storage, we regenerated the missing embeddings and out-of-fold
arrays. Of 366 v1 cells, 350 met the planned tolerance of 0.002 AUROC. Excluding ESM2-650M,
we recovered every T1 and T4 cell within 0.0004 and every T3 cell within 0.0012. Twelve exceptions
came from ESM2-650M, which we re-embedded with a different Transformers version. For genes with
0--10 publications the original and regenerated scores were T1 0.879 against 0.872 and
T4 0.597 against 0.592. The other four exceptions were MLP T2 scores for re-embedded text
(0.002--0.004). We retained the v1 values in the original comparison. Supplementary
Tables~\ref{tab:cluster} and \ref{tab:v11} use the regenerated ESM2 row where indicated.
Changing the family-grouped fold assignment (scikit-learn 1.9.1 versus 1.6.1) moves T2 cells by up to 0.010.

\paragraph{Analysis planning.}
The analyses followed sequential plans informed by earlier results (Supplementary Table~\ref{tab:rounds}).
For rounds 7 and 8, the plans and results entered the same Git commits. File timestamps support
the reported order but do not independently establish that planning preceded computation.
From round 9, plan commits preceded result commits. Round 14 also records a push to GitHub
before computation. We identify exploratory additions as post hoc, including the LLM replacement
and the human GO comparison with ESM2-650M. The subsequent release review checked labels and
saved aggregates without fitting additional models.

\paragraph{Release corrections.}\label{sec:releaseaudit}
The temporal archive originally stored a Boolean indicator of prior experimental annotation
in the publication-count field \texttt{exposure\_2023}. Applying the threshold of five to the
Boolean indicator included all 347 evaluation genes. Restoring integer counts from the released
gene table selects the intended 137. The original evaluation used those integer counts, so this
repair affects exported metadata. We checked that labels, folds and evaluation masks were preserved.

The historical GO evaluator assigned distinct ranks to tied scores. AUROC gives a positive and
a negative with equal scores half credit. The corrected implementation uses average ranks and
was checked against explicit pairwise comparisons. The error applies to GO scoring, including
the nearest-neighbour control. Binary endpoints, masked temporal scoring and T4 had separate,
tie-aware implementations. The original GO prediction arrays are unavailable, leaving the
correction to the reported GO contrasts and derived indices unknown. The term-frequency baseline
already used average ranks. Recomputing it recovered the per-gene means in Table~\ref{tab:exposure}
and gave pooled per-term AUROC 0.433. Within any one fold a term receives a constant score.
Across folds, a higher held-out prevalence lowers the complementary training frequency.

Seven non-exact symbol matches reused a GenePT key that was also the current symbol of another
gene in the study. Four duplicate-key groups spanned GO folds. Three secondary matches fell in
the 0--10 stratum, one contributed to its T1 evaluation and one was a T3 evaluation gene.
Resolving these mappings requires stable source identifiers and a refit, since a changed training
row can affect other genes' predictions. The release checks verified task inputs and recomputed
composite scores from saved results. They do not establish how corrected ties or mappings
would change a model's performance.

\begingroup
\small
\bibliography{references}
\endgroup

\clearpage
\appendix
\renewcommand{\theHfigure}{supp.\arabic{figure}}
\renewcommand{\theHtable}{supp.\arabic{table}}
\renewcommand{\thefigure}{S\arabic{figure}}
\renewcommand{\thetable}{S\arabic{table}}
\setcounter{figure}{0}
\setcounter{table}{0}
\section*{Supplementary information}

\subsection*{Benchmark profiles and sensitivity analyses}
\label{sec:suppbench}
\paragraph{Protein and text.}
Among genes with 0--10 publications, GB.Protein-16B (index 45.3, 44.5--46.0) scored above the
highest text-only entry (GenePT text-embedding-3-large, 43.0, paired difference $+2.2$, 1.5--3.0).
The other two sequence-only protein models also scored above all text-only entries
(ESM2-650M 44.4, ESM1b 43.9). The two GenePT releases, whose inputs include 31\% generated
summaries, fall from T1 ranks 2 and 7 to index ranks 7 and 11. All text-only representations
scored below the three protein models on T2 and T4 (Table~\ref{tab:bench}).

Three representations scored above GB.Protein-16B with a linear probe. One concatenated its
embedding with rich summary-filtered text (47.1, 46.3--47.8). The others were ProTrek-650M (47.0)
and ProtST-ESM1b (46.3), both trained on UniProt text. ProtST had the highest linear-probe score
on the annotation holdout (T3 0.836 against 0.780 for GB.Protein-16B). With the MLP, GB.Protein-16B
scored 54.3, above ProTrek (53.9) and ProtST (51.6). The mixture still ranked first
(Supplementary Table~\ref{tab:benchmlp}).

Removing UniProt text from the mixture lowered its index to 42.9, below GB.Protein-16B by 2.4
points ($-2.4$, $-2.9$ to $-1.8$). The text block alone fell from 42.8 to 37.9. Using the
regenerated ESM2-650M in the mixture gave 45.5, a gain of 1.8 over ESM2-650M (1.2--2.5) and
2.7 over rich text alone. Without UniProt the ESM2 mixture scored 41.2 (v1.1, Supplementary
Table~\ref{tab:v11}). Removing UniProt removes biological content and any benefit from shared
curation together.

\paragraph{Exposure and encoder choice.}
On all genes, rich summary-filtered text (47.8) and its name-masked version (46.9) both scored
above GB.Protein-16B (46.2). Text entries without generated summaries gained about 4--8 points
between the primary stratum and all genes (3.7 for symbols alone). The corresponding gains
were 1--3 for GenePT entries with generated text, 1.7--1.8 for text-aligned encoders, and at
most one point for sequence models trained without text (Fig.~\ref{fig:leaderboard}c). Among
sequence and expression models in the primary stratum, the protein models scored highest,
followed by RNA (GB.RNA-1.6B 39.9), single-cell (36.6, 34.7) and DNA (GB.DNA-300M 31.0).
UniProt function text alone (42.5) matches rich summary-filtered text on T1 (0.916 vs 0.915)
and trails all three protein models on T2 and T4. In Track F the five neural encoders gave
removal contrasts of 0.078--0.092 and TF-IDF 0.043. The six summary-filtered T1 scores differ
by at most 0.013, and their indices lie within 2.3 points of one another.

\begin{figure}[p]
  \centering
  \includegraphics[width=\textwidth]{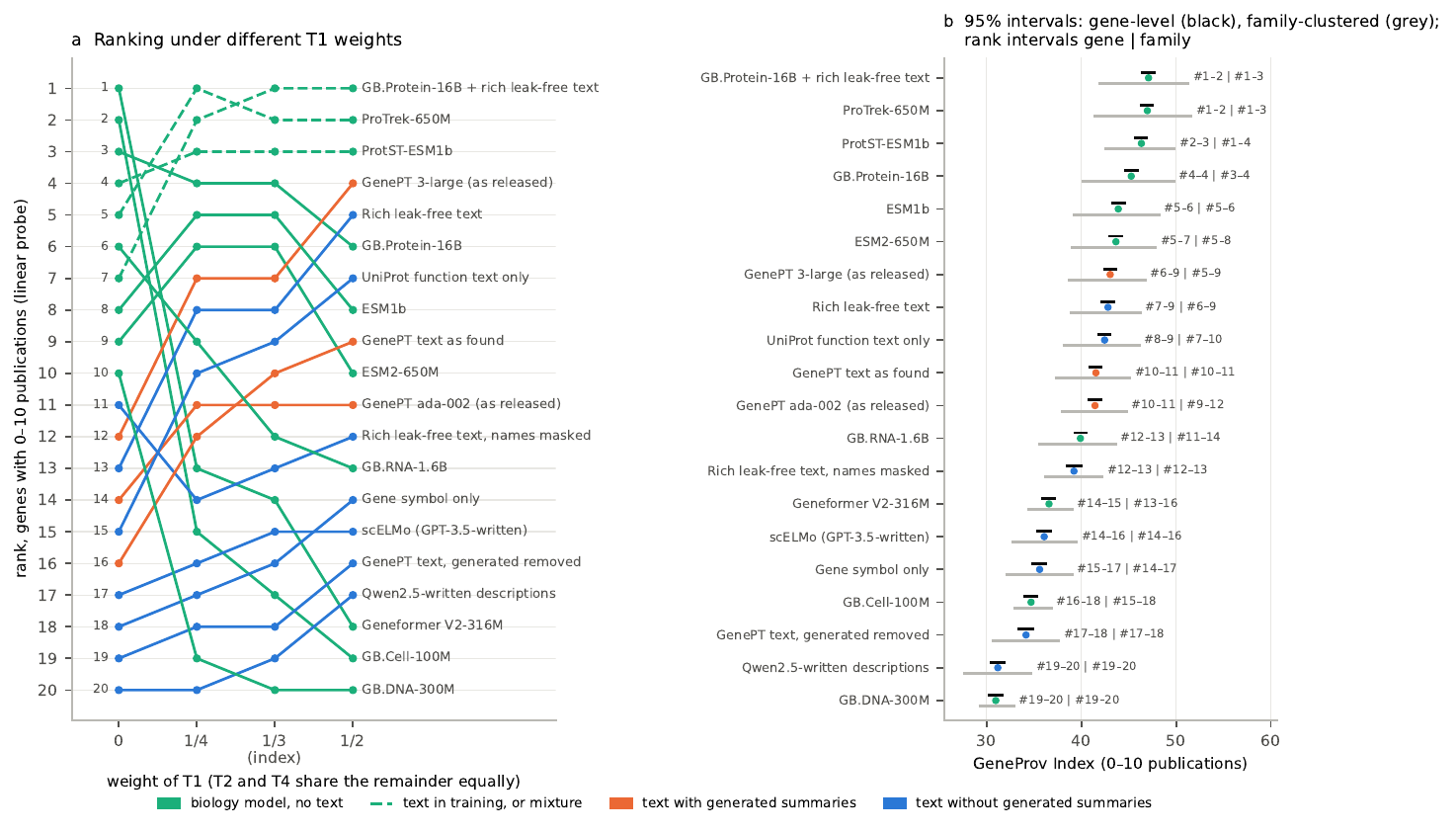}
  \caption{\textbf{Rank sensitivity to task weights and family clustering.} Here “leak-free” denotes summary-filtered text. (a) Rank of every
  Track O representation (linear probe, genes with 0--10 publications) when T1 receives weight 0, 1/4, 1/3 (the
  GeneProv Index) or 1/2 and T2 and T4 share the remainder equally. (b) The headline index with gene-level (black)
  and family-clustered (grey) 95\% bootstrap intervals, and the rank interval under each; numbers in Supplementary
  Table~\ref{tab:cluster}.}
  \label{fig:weights}
\end{figure}

\paragraph{Task weighting and probe.}
The rank correlation between the linear and MLP indices was $\rho = 0.96$. With the MLP, the
mixture leads at 58.2 and GB.Protein-16B is the best single row at 54.3. Replacing mean skill
with mean rank gave a similar ordering ($\rho = 0.85$). Dropping T1 had a much larger effect
($\rho = 0.43$ for the original T2/T4-only index). The single-cell models ranked first, and all
text-only entries scored below every sequence and expression model. Giving T1 a weight of 1/4 or
1/2 preserved much of the original ordering (Spearman 0.98 and 0.94). At 1/4 all three sequence-only
protein models remained above every text-only model. At 1/2 two text models passed ESM1b and
ESM2-650M. At zero weight the single-cell models lead (0.44 in the regenerated weight sweep,
Supplementary Fig.~\ref{fig:weights}).

\paragraph{Label coverage and family clustering.}
Holding the T2 labels fixed by excluding ClinGen from the all-gene average raised T2 by up to
0.018 and moved the all-gene ranks by at most two positions (Spearman 0.99). With fixed labels,
the sequence models other than DNA gained 1.1--2.0 index points between the primary stratum and
all genes. Family resampling widened the score intervals (GB.Protein-16B 40.1--49.9 against
44.5--46.0) but shifted the endpoints of rank intervals by at most two positions. GB.Protein-16B
ranks 3--4 under family resampling. One family accounted for a substantial share of the sample. In all,
531 of the 7,120 genes in the primary stratum are C2H2 zinc fingers. The main summary-removal
contrast had a family-clustered interval 0.070--0.086 (gene-level 0.072--0.084).
Limiting comparison to the 18,308 genes covered by all entries (6,354 with 0--10 publications)
also moved ranks by at most two positions (Spearman 0.99, Supplementary
Tables~\ref{tab:cluster} and \ref{tab:coverage}).

\paragraph{Co-essential partners among the first ten candidates.}
We also counted how many of a gene's ten co-essential partners appeared in its first ten candidates.
The resulting model ranking correlated weakly with the AUROC ranking (Spearman 0.03 with the AUROC
order, post hoc). GB.RNA-1.6B has the highest T4 AUROC (0.628) but retrieves 0.8\% of the
partners in its top ten. The single-cell and DNA models retrieved 0.4--0.5\%, the text-only
models 1.4--2.0\%, the protein models 1.5--1.7\%, and the mixture 2.4\%.

\begin{figure}[p]
  \centering
  \includegraphics[width=\textwidth]{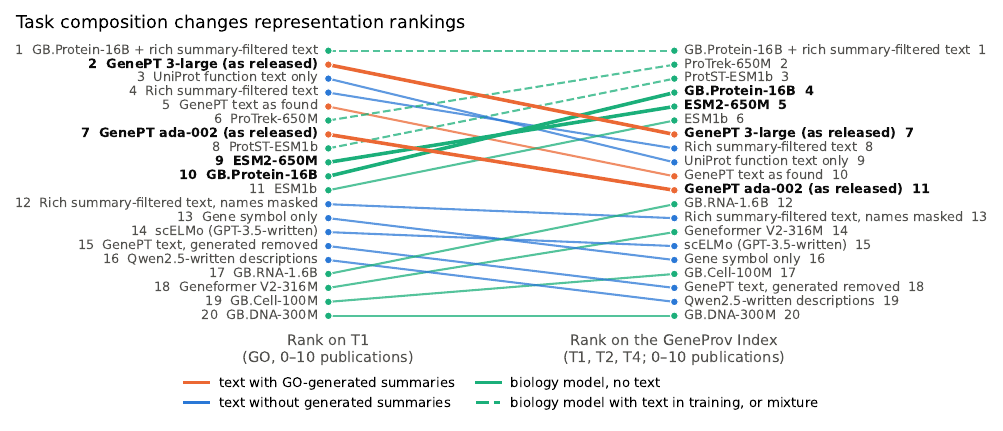}
  \caption{\textbf{Task composition changes representation rankings.} The same twenty Track O representations
  ranked on GO (T1) and on the equal-weight T1/T2/T4 index for genes with 0--10 publications, linear probe.
  Inputs are held fixed between the two rankings; Track O includes representations containing generated summaries.
  Differences in rank reflect the choice of tasks.}
  \label{fig:taskranks}
\end{figure}

\begin{figure}[htbp]
  \centering
  \includegraphics[width=0.82\textwidth]{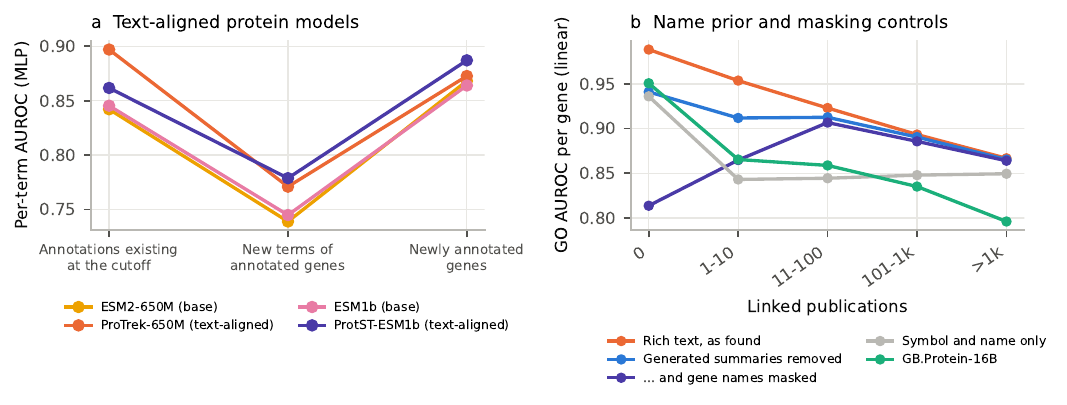}
  \caption{\textbf{Supplementary controls.} (a) Text-aligned protein models against their base encoders on annotations
  that existed at the cutoff, new terms of annotated genes, and genes newly represented in the selected benchmark
  labels (MLP probe, per-term AUROC).
  (b) Name prior and masking controls on the rich input, by literature exposure (per-gene GO AUROC, linear probe).}
  \label{fig:supp}
\end{figure}

\begin{table}[htbp]
  \centering
  \caption{\textbf{Per-term GO AUROC within each exposure bin.} GenePT's 2023 text embedded with Qwen3-Embedding-8B,
  standard labels; AUROC of each term over the genes of one bin only, averaged over terms with at least five positive
  and twenty negative genes in that bin (number of terms in the header). Replacement by LLM-written text is post hoc.}
  \label{tab:perterm}
  \small\setlength{\tabcolsep}{3.8pt}
  \begin{tabular}{@{}llccccc@{}}
    \toprule
    Input & Probe & 0 (152) & 1--10 (2,750) & 11--100 (4,798) & 101--1000 (4,618) & $>$1000 (2,789) \\
    \midrule
    As found              & linear & 0.949 & 0.871 & 0.818 & 0.781 & 0.717 \\
    Generated removed     & linear & 0.922 & 0.780 & 0.798 & 0.784 & 0.724 \\
    Replaced by LLM text  & linear & 0.929 & 0.786 & 0.797 & 0.782 & 0.716 \\
    As found              & MLP    & 0.968 & 0.908 & 0.868 & 0.847 & 0.792 \\
    Generated removed     & MLP    & 0.941 & 0.832 & 0.847 & 0.846 & 0.789 \\
    Replaced by LLM text  & MLP    & 0.950 & 0.829 & 0.839 & 0.844 & 0.787 \\
    \bottomrule
  \end{tabular}
\end{table}

\begin{table}[htbp]
  \centering
  \caption{\textbf{Per-gene GO AUROC by literature exposure, MLP probe.} As Table~\ref{tab:exposure} with the
  one-hidden-layer MLP probe; the last column is per-term AUROC over all genes. Rows marked $\dagger$ were added after
  the analysis plan; the symbol-only input was run with the linear probe only.}
  \label{tab:mlp}
  \small\setlength{\tabcolsep}{3.9pt}
  \begin{tabular}{@{}lccccccc@{}}
    \toprule
    Input & 0 & 1--10 & 11--100 & 101--1000 & $>$1000 & all & per term \\
    \midrule
    GenePT text, as found                         & 0.986 & 0.952 & 0.928 & 0.903 & 0.871 & 0.930 & 0.902 \\
    GenePT text, generated summaries removed      & 0.972 & 0.918 & 0.923 & 0.912 & 0.881 & 0.920 & 0.872 \\
    GenePT text, replaced by LLM-written text$^\dagger$ & 0.968 & 0.906 & 0.910 & 0.903 & 0.869 & 0.909 & 0.876 \\
    Rich text, as found                           & 0.993 & 0.978 & 0.957 & 0.929 & 0.895 & 0.956 & 0.932 \\
    Rich text, generated summaries removed        & 0.982 & 0.965 & 0.952 & 0.926 & 0.894 & 0.950 & 0.933 \\
    Rich text, removed and gene names masked      & 0.930 & 0.948 & 0.950 & 0.926 & 0.892 & 0.943 & 0.905 \\
    GB.Protein-16B                                & 0.979 & 0.940 & 0.926 & 0.903 & 0.874 & 0.926 & 0.853 \\
    ESM2-650M$^\dagger$                            & 0.982 & 0.942 & 0.929 & 0.909 & 0.878 & 0.929 & 0.853 \\
    \bottomrule
  \end{tabular}
\end{table}

\begin{table}[htbp]
  \centering
  \caption{\textbf{UniProt function text in place of the NCBI summary.} Per-gene GO AUROC, linear probe, standard
  labels, restricted to the genes whose summary was replaced. Top: generated-summary genes with a UniProt function
  text (4,591); bottom: RefSeq-summary genes with one (11,074). Differences carry 95\% bootstrap intervals.}
  \label{tab:uniprot}
  \footnotesize\setlength{\tabcolsep}{2.8pt}
  \begin{tabular}{@{}lrccccl@{}}
    \toprule
    Genes / bin & $n$ & As found & UniProt text & Removed & As found $-$ UniProt & As found $-$ removed \\
    \midrule
    Generated, 0        &    95 & 0.948 & 0.887 & 0.887 & $+0.061$ (0.032 to 0.095)   & $+0.061$ \\
    Generated, 1--10    & 1,938 & 0.909 & 0.905 & 0.756 & $+0.004$ ($-0.004$ to 0.012) & $+0.153$ \\
    Generated, 11--100  & 1,712 & 0.902 & 0.895 & 0.774 & $+0.007$ ($-0.000$ to 0.014) & $+0.128$ \\
    Generated, 101--1000 &  197 & 0.874 & 0.854 & 0.789 & $+0.020$ (0.001 to 0.041)   & $+0.086$ \\
    \midrule
    RefSeq, 1--10       & 1,353 & 0.901 & 0.933 &       & $-0.032$ ($-0.040$ to $-0.024$) & \\
    RefSeq, 11--100     & 5,763 & 0.881 & 0.911 &       & $-0.031$ ($-0.034$ to $-0.027$) & \\
    RefSeq, 101--1000   & 2,998 & 0.862 & 0.889 &       & $-0.026$ ($-0.030$ to $-0.023$) & \\
    RefSeq, $>$1000     &   404 & 0.838 & 0.859 &       & $-0.020$ ($-0.029$ to $-0.012$) & \\
    \bottomrule
  \end{tabular}
\end{table}

\begin{figure}[htbp]
  \centering
  \includegraphics[width=0.82\textwidth]{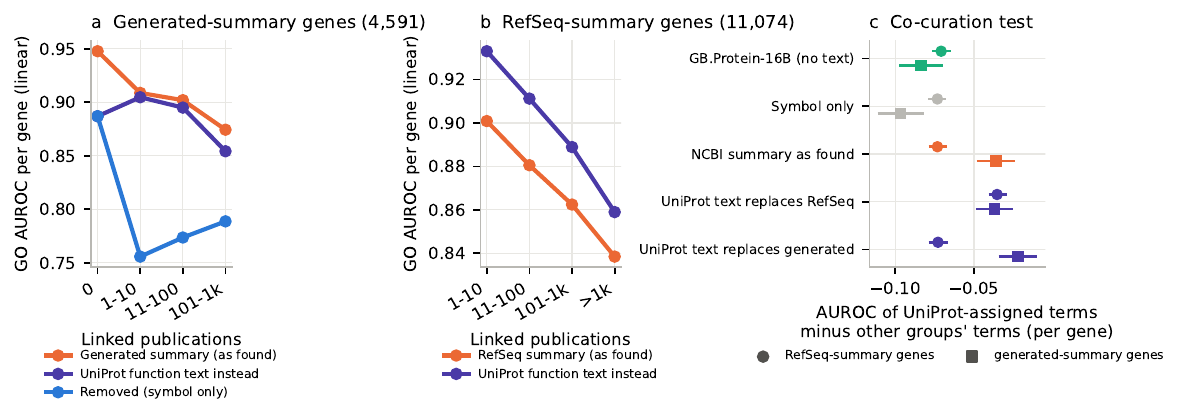}
  \caption{\textbf{Curated UniProt text behaves like the generated summaries.} (a) Generated-summary genes with a
  UniProt function text: per-gene GO AUROC with the generated summary, with the UniProt text in its place, and with
  the symbol alone. (b) RefSeq-summary genes: with the RefSeq summary and with the UniProt text in its place.
  (c) Co-curation test: per-gene AUROC of UniProt-assigned direct terms minus that of terms assigned by other groups,
  each against the gene's negatives, for five inputs; circles RefSeq-summary genes, squares generated-summary genes;
  95\% bootstrap intervals.}
  \label{fig:uniprot}
\end{figure}

\begin{table}[htbp]
  \centering
  \caption{\textbf{Generated NCBI Gene summaries by organism and literature exposure.} Share (\%) of protein-coding
  genes (NCBI gene\_info, October 2026) whose NCBI Gene summary is provided by the Alliance of Genome Resources.
  Exposure: PubMed articles linked to the gene in gene2pubmed, counting articles linked to at most 20 genes of the same
  organism. Human counts here use NCBI's protein-coding set; the main text uses the HGNC set (32.1\%). The NCBI set includes
about 1,300 further genes without publications that have no summary of any kind (55\% of that bin), which lowers the
zero-bin share to 29\% against 56\% for the HGNC set in Fig.~\ref{fig:prevalence}a.}
  \label{tab:organisms}
  \small
  \begin{tabular}{@{}lrrrrrrr@{}}
    \toprule
    Organism & Genes & All & 0 & 1--10 & 11--100 & 101--1000 & RefSeq summary \\
    \midrule
    Human     & 20,606 & 30.0 & 29.0 & 56.6 & 22.1 &   6.4 & 58.2 \\
    Mouse     & 26,365 & 69.6 & 32.7 & 90.0 & 89.6 &  81.6 & 10.0 \\
    Rat       & 25,140 & 75.8 & 42.7 & 97.6 & 97.6 &  95.2 &  5.5 \\
    Zebrafish & 34,930 & 61.5 & 39.2 & 96.4 & 99.0 &  98.2 &  0.2 \\
    Fly       & 13,986 & 86.0 & 59.2 & 92.6 & 99.7 & 100.0 &  0.0 \\
    Worm      & 19,983 & 63.8 & 47.5 & 93.9 & 99.9 & 100.0 &  0.0 \\
    Yeast     &  6,029 & 95.1 & 58.9 & 94.8 & 100.0 & 100.0 & 0.0 \\
    \bottomrule
  \end{tabular}
\end{table}

\input{table_bench_v1_mlp}

\input{table_r14_v11}

\input{table_r14_zero}

\input{table_r14_cluster}

\input{table_r14_coverage}

\input{table_r14_models}

\input{table_r14_rounds}

\end{document}

%% file: table_bench_v1.tex
\begin{table}[t]
  \centering
  \caption{\textbf{GeneProv v1 leaderboard.} Track O (open representation), linear probe (T4 has no probe), sorted by
  the headline GeneProv Index: the mean over T1 (GO), T2 (non-GO labels, family-grouped folds) and T4
  (co-essential partner retrieval) of the chance-corrected skill $100\,(2\,\mathrm{AUROC}-1)$, for genes with 0--10
  publications, with its 95\% paired bootstrap interval over genes; \# is the rank with its 95\% rank interval.
  T3 (for 0--10: genes with 0--5 publications at the cutoff; experimental GO annotations added after 1 January 2023; reference literature-only 2023 text, MPNet: 0.732 for 0--5, 0.800 for all) is
  reported beside the index for representations eligible under the input-date rule in Methods. Track O declares
  text provenance and includes released embeddings that contain generated summaries. text: share (\%) of the representation's genes
  whose embedded text is generated from GO; \emph{train}: no text embedded but UniProt function text used in
  training; \emph{mix}: declared mixture. $\ddagger$: embedded with Qwen3-Embedding-8B. Track F (fixed text):
  embedders scored on corpus L (GenePT's 2023 text, generated summaries removed), with the same index and $S$ = per-gene
  GO AUROC on corpus A (as found) minus corpus L for genes with 1--10 publications (95\% interval). MLP probe in
  Supplementary Table~\ref{tab:benchmlp}; task definitions in Methods.}
  \label{tab:bench}
  \scriptsize\setlength{\tabcolsep}{1.9pt}
  \begin{tabular}{@{}r>{\raggedright\arraybackslash}p{0.24\linewidth}rlccccccccc@{}}
    \toprule
    & & & \multicolumn{2}{c}{Index} & \multicolumn{4}{c}{Genes with 0--10 publications} & \multicolumn{4}{c}{All genes} \\
    \cmidrule(lr){4-5} \cmidrule(lr){6-9} \cmidrule(lr){10-13}
    \# (95\%) & Representation & text & 0--10 (95\%) & all & T1 & T2 & T4 & T3 & T1 & T2 & T4 & T3 \\
    \midrule
      1 (1--2) & GB.Protein-16B $\oplus$ rich summary-filtered text$^{\ddagger}$ &   mix & 47.1 (46.3--47.8) & 50.4 & 0.921 & 0.658 & 0.627 & -- & 0.908 & 0.720 & 0.628 & -- \\
      2 (1--2) & ProTrek-650M (text-aligned)                          & train & 47.0 (46.2--47.7) & 48.8 & 0.904 & 0.698 & 0.603 & -- & 0.891 & 0.749 & 0.592 & -- \\
      3 (2--3) & ProtST-ESM1b (text-aligned)                          & train & 46.3 (45.6--47.0) & 48.0 & 0.892 & 0.729 & 0.574 & 0.836 & 0.878 & 0.773 & 0.570 & 0.865 \\
             4 & GB.Protein-16B                                       &    -- & 45.3 (44.5--46.0) & 46.2 & 0.875 & 0.676 & 0.627 & 0.780 & 0.857 & 0.719 & 0.616 & 0.833 \\
      5 (5--6) & ESM2-650M                                            &    -- & 44.4 (43.7--45.1) & 44.9 & 0.879 & 0.690 & 0.597 & 0.777 & 0.860 & 0.729 & 0.585 & 0.839 \\
      6 (5--6) & ESM1b                                                &    -- & 43.9 (43.1--44.7) & 44.5 & 0.873 & 0.674 & 0.612 & 0.769 & 0.852 & 0.714 & 0.601 & 0.825 \\
      7 (7--9) & GenePT text-emb.-3-large (as released)               &    31 & 43.0 (42.3--43.8) & 46.0 & 0.919 & 0.656 & 0.571 & -- & 0.898 & 0.713 & 0.580 & -- \\
      8 (7--9) & Rich summary-filtered text$^{\ddagger}$              &     0 & 42.8 (42.0--43.6) & 47.8 & 0.915 & 0.651 & 0.576 & -- & 0.908 & 0.722 & 0.587 & -- \\
      9 (8--9) & UniProt function text only$^{\ddagger}$              &     0 & 42.5 (41.7--43.2) & 46.6 & 0.916 & 0.647 & 0.574 & -- & 0.907 & 0.706 & 0.585 & -- \\
    10 (10--11) & GenePT text as found$^{\ddagger}$                    &    30 & 41.5 (40.7--42.2) & 44.4 & 0.904 & 0.653 & 0.565 & -- & 0.883 & 0.713 & 0.571 & -- \\
    11 (10--11) & GenePT ada-002 (NCBI text, as released)              &    31 & 41.4 (40.7--42.2) & 42.5 & 0.899 & 0.661 & 0.562 & 0.700 & 0.872 & 0.701 & 0.565 & 0.784 \\
    12 (12--13) & GB.RNA-1.6B                                          &    -- & 39.9 (39.2--40.6) & 40.4 & 0.800 & 0.671 & 0.628 & 0.753 & 0.785 & 0.709 & 0.613 & 0.782 \\
    13 (12--13) & Rich summary-filtered text, names masked$^{\ddagger}$ &     0 & 39.2 (38.4--40.1) & 46.9 & 0.859 & 0.655 & 0.575 & -- & 0.888 & 0.729 & 0.587 & -- \\
    14 (14--15) & Geneformer V2-316M                                   &    -- & 36.6 (35.8--37.3) & 37.7 & 0.728 & 0.733 & 0.588 & 0.666 & 0.731 & 0.746 & 0.590 & 0.690 \\
    15 (14--16) & scELMo (GPT-3.5-written descriptions)                &     0 & 36.1 (35.2--37.0) & 42.3 & 0.839 & 0.643 & 0.559 & 0.733 & 0.855 & 0.707 & 0.572 & 0.817 \\
    16 (15--17) & Gene symbol only$^{\ddagger}$                        &     0 & 35.6 (34.7--36.4) & 39.3 & 0.854 & 0.623 & 0.557 & -- & 0.848 & 0.676 & 0.566 & -- \\
    17 (16--18) & GB.Cell-100M                                         &    -- & 34.7 (33.9--35.5) & 37.6 & 0.708 & 0.733 & 0.580 & 0.679 & 0.729 & 0.757 & 0.578 & 0.719 \\
    18 (17--18) & GenePT text, generated summaries removed$^{\ddagger}$ &     0 & 34.2 (33.3--35.1) & 40.7 & 0.832 & 0.624 & 0.556 & -- & 0.849 & 0.708 & 0.553 & -- \\
    19 (19--20) & Qwen2.5-7B-written descriptions$^{\ddagger}$         &     0 & 31.2 (30.3--32.0) & 35.7 & 0.819 & 0.600 & 0.549 & -- & 0.820 & 0.658 & 0.557 & -- \\
    20 (19--20) & GB.DNA-300M                                          &    -- & 31.0 (30.1--31.8) & 29.3 & 0.704 & 0.678 & 0.582 & 0.645 & 0.683 & 0.685 & 0.571 & 0.661 \\
    \bottomrule
  \end{tabular}

  \vspace{0.6em}
  \begin{tabular}{@{}lcccccccl@{}}
    \toprule
    Track F embedder & Index 0--10 & Index all & T1 0--10 & T2 0--10 & T4 0--10 & T1 all & $S$ (1--10) \\
    \midrule
    pubmedbert-base-embeddings   & 34.6 & 41.4 & 0.822 & 0.641 & 0.557 & 0.846 & 0.090 (0.083--0.096) \\
    Qwen3-Embedding-8B           & 34.2 & 40.7 & 0.832 & 0.624 & 0.556 & 0.849 & 0.078 (0.072--0.084) \\
    all-mpnet-base-v2            & 33.3 & 39.1 & 0.820 & 0.626 & 0.555 & 0.835 & 0.088 (0.082--0.094) \\
    Qwen3-Embedding-0.6B         & 32.9 & 38.7 & 0.822 & 0.621 & 0.552 & 0.838 & 0.088 (0.082--0.093) \\
    TF-IDF (1,024-d SVD)         & 32.5 & 35.0 & 0.830 & 0.613 & 0.544 & 0.819 & 0.043 (0.038--0.048) \\
    bge-large-en-v1.5            & 32.3 & 39.1 & 0.821 & 0.611 & 0.552 & 0.839 & 0.092 (0.086--0.098) \\
    \bottomrule
  \end{tabular}
\end{table}

%% file: table_bench_v1_mlp.tex
\begin{table}[h]
  \centering
  \caption{\textbf{GeneProv v1, MLP probe.} Table~\ref{tab:bench} with the one-hidden-layer MLP probe (T4 is
  probe-free and identical); sorted by the headline index under this probe. Reference for T3 (literature-only 2023
  text, MPNet): 0.877 for 0--5, 0.903 for all.}
  \label{tab:benchmlp}
  \scriptsize\setlength{\tabcolsep}{1.9pt}
  \begin{tabular}{@{}r>{\raggedright\arraybackslash}p{0.24\linewidth}rlccccccccc@{}}
    \toprule
    & & & \multicolumn{2}{c}{Index} & \multicolumn{4}{c}{Genes with 0--10 publications} & \multicolumn{4}{c}{All genes} \\
    \cmidrule(lr){4-5} \cmidrule(lr){6-9} \cmidrule(lr){10-13}
    \# (95\%) & Representation & text & 0--10 & all & T1 & T2 & T4 & T3 & T1 & T2 & T4 & T3 \\
    \midrule
             1 & GB.Protein-16B $\oplus$ rich summary-filtered text$^{\ddagger}$ &   mix & 58.2 & 59.5 & 0.970 & 0.777 & 0.627 & -- & 0.952 & 0.812 & 0.628 & -- \\
      2 (2--3) & GB.Protein-16B                                       &    -- & 54.3 & 53.6 & 0.945 & 0.743 & 0.627 & 0.917 & 0.926 & 0.763 & 0.616 & 0.928 \\
      3 (2--3) & ProTrek-650M (text-aligned)                          & train & 53.9 & 54.0 & 0.960 & 0.747 & 0.603 & -- & 0.943 & 0.775 & 0.592 & -- \\
             4 & ESM1b                                                &    -- & 53.2 & 52.2 & 0.945 & 0.742 & 0.612 & 0.916 & 0.928 & 0.755 & 0.601 & 0.928 \\
      5 (5--7) & ESM2-650M                                            &    -- & 52.1 & 51.4 & 0.946 & 0.739 & 0.597 & 0.914 & 0.929 & 0.757 & 0.585 & 0.930 \\
      6 (5--7) & Rich summary-filtered text$^{\ddagger}$              &     0 & 52.1 & 55.1 & 0.967 & 0.739 & 0.576 & -- & 0.950 & 0.790 & 0.587 & -- \\
      7 (6--8) & UniProt function text only$^{\ddagger}$              &     0 & 51.6 & 54.0 & 0.967 & 0.733 & 0.574 & -- & 0.949 & 0.776 & 0.585 & -- \\
      8 (6--8) & ProtST-ESM1b (text-aligned)                          & train & 51.6 & 51.8 & 0.949 & 0.750 & 0.574 & 0.920 & 0.936 & 0.770 & 0.570 & 0.929 \\
             9 & GenePT text-emb.-3-large (as released)               &    31 & 50.9 & 52.6 & 0.964 & 0.728 & 0.571 & -- & 0.942 & 0.767 & 0.580 & -- \\
            10 & GenePT text as found$^{\ddagger}$                    &    30 & 50.0 & 51.8 & 0.956 & 0.730 & 0.565 & -- & 0.930 & 0.776 & 0.571 & -- \\
    11 (11--12) & Rich summary-filtered text, names masked$^{\ddagger}$ &     0 & 47.9 & 53.5 & 0.946 & 0.698 & 0.575 & -- & 0.943 & 0.773 & 0.587 & -- \\
    12 (11--13) & GenePT ada-002 (NCBI text, as released)              &    31 & 47.3 & 47.9 & 0.954 & 0.694 & 0.562 & 0.905 & 0.929 & 0.725 & 0.565 & 0.916 \\
    13 (11--13) & GB.RNA-1.6B                                          &    -- & 47.3 & 46.0 & 0.887 & 0.694 & 0.628 & 0.873 & 0.867 & 0.709 & 0.613 & 0.870 \\
    14 (14--15) & Geneformer V2-316M                                   &    -- & 45.9 & 45.4 & 0.864 & 0.738 & 0.588 & 0.854 & 0.843 & 0.748 & 0.590 & 0.847 \\
    15 (14--15) & Gene symbol only$^{\ddagger}$                        &     0 & 45.9 & 48.6 & 0.933 & 0.698 & 0.557 & -- & 0.918 & 0.745 & 0.566 & -- \\
    16 (16--17) & GB.Cell-100M                                         &    -- & 44.9 & 45.6 & 0.866 & 0.727 & 0.580 & 0.857 & 0.854 & 0.752 & 0.578 & 0.848 \\
    17 (16--17) & scELMo (GPT-3.5-written descriptions)                &     0 & 44.8 & 48.6 & 0.932 & 0.680 & 0.559 & 0.900 & 0.926 & 0.731 & 0.572 & 0.924 \\
    18 (17--18) & GenePT text, generated summaries removed$^{\ddagger}$ &     0 & 43.9 & 48.6 & 0.925 & 0.679 & 0.556 & -- & 0.920 & 0.756 & 0.553 & -- \\
            19 & Qwen2.5-7B-written descriptions$^{\ddagger}$         &     0 & 40.4 & 44.0 & 0.906 & 0.650 & 0.549 & -- & 0.893 & 0.710 & 0.557 & -- \\
            20 & GB.DNA-300M                                          &    -- & 38.9 & 36.3 & 0.841 & 0.661 & 0.582 & 0.805 & 0.814 & 0.660 & 0.571 & 0.786 \\
    \bottomrule
  \end{tabular}

  \vspace{0.6em}
  \begin{tabular}{@{}lcccccccl@{}}
    \toprule
    Track F embedder & Index 0--10 & Index all & T1 0--10 & T2 0--10 & T4 0--10 & T1 all & $S$ (1--10) \\
    \midrule
    Qwen3-Embedding-8B           & 43.9 & 48.6 & 0.925 & 0.679 & 0.556 & 0.920 & 0.033 (0.030--0.037) \\
    pubmedbert-base-embeddings   & 41.9 & 46.2 & 0.924 & 0.649 & 0.557 & 0.921 & 0.040 (0.038--0.043) \\
    all-mpnet-base-v2            & 41.0 & 45.0 & 0.921 & 0.640 & 0.555 & 0.915 & 0.038 (0.035--0.041) \\
    Qwen3-Embedding-0.6B         & 41.0 & 44.3 & 0.920 & 0.644 & 0.552 & 0.912 & 0.038 (0.035--0.041) \\
    bge-large-en-v1.5            & 41.0 & 45.5 & 0.920 & 0.643 & 0.552 & 0.915 & 0.040 (0.037--0.043) \\
    TF-IDF (1,024-d SVD)         & 39.3 & 39.4 & 0.920 & 0.625 & 0.544 & 0.901 & 0.021 (0.019--0.023) \\
    \bottomrule
  \end{tabular}
\end{table}

%% file: table_r14_v11.tex

\begin{table}[p]
  \centering
  \caption{\textbf{Rows appended in v1.1 (round 14).} Linear probe (MLP in brackets for the index), genes with 0--10 publications and all genes, with the v1 rows they are compared with. Top: the mixture control without UniProt function text and the ESM2-650M mixtures; the rich summary-filtered text without UniProt function text is ``Gene SYMBOL (name). RefSeq summary''. Bottom: T3 uses the gene's GO annotation vector at the cutoff (standard rule, non-IEA, 4,894 terms) alone and concatenated with GB.Protein-16B, scored on genes with 0--5 publications at the cutoff and on all evaluation genes, primary targets and the novel-pair variant; these rows take no other task. Differences carry 95\% paired bootstrap intervals over genes. The ESM2 comparator uses regenerated embeddings (Methods). Version labels identify when each representation entered the benchmark.}
  \label{tab:v11}
  \scriptsize\setlength{\tabcolsep}{1.7pt}
  \resizebox{\textwidth}{!}{%
  \begin{tabular}{@{}P{0.225\linewidth}ccccccccc@{}}
    \toprule
    & & \multicolumn{4}{c}{Genes with 0--10 publications} & \multicolumn{4}{c}{All genes} \\ \cmidrule(lr){3-6} \cmidrule(lr){7-10}
    Representation & version & Index (95\%) [MLP] & T1 & T2 & T4 & Index [MLP] & T1 & T2 & T4 \\
    \midrule
    GB.Protein-16B & v1 & 45.3 (44.5--46.0) [54.3] & 0.875 & 0.676 & 0.627 & 46.2 [53.6] & 0.857 & 0.719 & 0.616 \\
    ESM2-650M & v1 & 43.6 (42.9--44.4) [51.1] & 0.872 & 0.690 & 0.592 & 44.1 [50.0] & 0.852 & 0.728 & 0.582 \\
    Rich summary-filtered text & v1 & 42.8 (42.0--43.6) [52.1] & 0.915 & 0.651 & 0.576 & 47.8 [55.1] & 0.908 & 0.722 & 0.587 \\
    UniProt function text only & v1 & 42.5 (41.7--43.2) [51.6] & 0.916 & 0.647 & 0.574 & 46.6 [54.0] & 0.907 & 0.706 & 0.585 \\
    GB.Protein-16B $+$ rich summary-filtered text & v1 & 47.1 (46.3--47.8) [58.2] & 0.921 & 0.658 & 0.627 & 50.4 [59.5] & 0.908 & 0.720 & 0.628 \\
    Rich summary-filtered text without UniProt text & v1.1 & 37.9 (37.0--38.6) [46.9] & 0.860 & 0.640 & 0.568 & 42.6 [50.3] & 0.862 & 0.709 & 0.567 \\
    GB.Protein-16B $+$ text without UniProt & v1.1 & 42.9 (42.1--43.6) [54.9] & 0.874 & 0.650 & 0.619 & 46.1 [56.4] & 0.867 & 0.711 & 0.613 \\
    ESM2-650M $+$ rich summary-filtered text & v1.1 & 45.5 (44.8--46.2) [55.4] & 0.921 & 0.661 & 0.600 & 49.3 [57.6] & 0.909 & 0.723 & 0.607 \\
    ESM2-650M $+$ text without UniProt & v1.1 & 41.2 (40.5--41.9) [51.7] & 0.873 & 0.653 & 0.592 & 44.7 [54.0] & 0.868 & 0.714 & 0.590 \\
    \midrule
    \multicolumn{10}{@{}P{\linewidth}@{}}{Headline index differences: mixture without UniProt text minus GB.Protein-16B $-2.4$ (-2.9 to -1.8); v1 mixture minus mixture without UniProt text $+4.2$ (+3.7 to +4.6);} \\
    \multicolumn{10}{@{}P{\linewidth}@{}}{ESM2-650M mixture minus ESM2-650M $+1.8$ (+1.2 to +2.5), minus rich summary-filtered text $+2.7$ (+2.3 to +3.0); ESM2-650M mixture minus its version without UniProt text $+4.3$ (+3.8 to +4.8).} \\
    \midrule
    \multicolumn{10}{@{}P{\linewidth}@{}}{\emph{T3: per-gene AUROC on experimental annotations added after the 2023 cutoff; linear [MLP]}} \\
    Representation & version & \multicolumn{2}{c}{0--5, primary} & \multicolumn{2}{c}{0--5, novel pairs} & \multicolumn{2}{c}{all, primary} & \multicolumn{2}{c}{all, novel pairs} \\
    GB.Protein-16B & v1 & \multicolumn{2}{c}{0.780 [0.917]} & \multicolumn{2}{c}{0.743 [0.895]} & \multicolumn{2}{c}{0.833 [0.928]} & \multicolumn{2}{c}{0.765 [0.894]} \\
    ESM2-650M & v1 & \multicolumn{2}{c}{0.779 [0.913]} & \multicolumn{2}{c}{0.742 [0.895]} & \multicolumn{2}{c}{0.835 [0.927]} & \multicolumn{2}{c}{0.770 [0.897]} \\
    GenePT ada-002 (as released) & v1 & \multicolumn{2}{c}{0.700 [0.905]} & \multicolumn{2}{c}{0.641 [0.879]} & \multicolumn{2}{c}{0.784 [0.916]} & \multicolumn{2}{c}{0.706 [0.883]} \\
    GO-2023 annotation vector & v1.1 & \multicolumn{2}{c}{0.493 [0.889]} & \multicolumn{2}{c}{0.350 [0.860]} & \multicolumn{2}{c}{0.603 [0.910]} & \multicolumn{2}{c}{0.399 [0.860]} \\
    GB.Protein-16B $+$ GO-2023 vector & v1.1 & \multicolumn{2}{c}{0.540 [0.898]} & \multicolumn{2}{c}{0.414 [0.867]} & \multicolumn{2}{c}{0.633 [0.918]} & \multicolumn{2}{c}{0.445 [0.869]} \\
    \midrule
    \multicolumn{10}{@{}P{\linewidth}@{}}{Annotation vector minus frozen 2023 text (MPNet), 0--5, primary: linear $-0.240$ (-0.303 to -0.182, $n$ = 137), MLP $+0.012$ (-0.012 to +0.034, $n$ = 137);} \\
    \multicolumn{10}{@{}P{\linewidth}@{}}{GB.Protein-16B minus annotation vector: linear $+0.288$ (+0.229 to +0.350, $n$ = 137), MLP $+0.028$ (+0.005 to +0.055, $n$ = 137); novel pairs, MLP $+0.035$ (+0.000 to +0.072, $n$ = 118);} \\
    \multicolumn{10}{@{}P{\linewidth}@{}}{GB.Protein-16B $\oplus$ vector minus GB.Protein-16B: linear $-0.240$ (-0.298 to -0.184, $n$ = 137), MLP $-0.020$ (-0.040 to +0.000, $n$ = 137); all genes, MLP $-0.010$ (-0.022 to +0.002, $n$ = 347).} \\
    \multicolumn{10}{@{}P{\linewidth}@{}}{Primary targets: 3,224 gene--term pairs over 347 genes; 34.6\% already carried a non-experimental annotation of the same term in 2023; novel pairs: 2,107.} \\
    \bottomrule
  \end{tabular}
  }
\end{table}

%% file: table_r14_zero.tex

\begin{table}[p]
  \centering
  \caption{\textbf{The zero-literature bin beside the 1--10 bin and the pooled headline stratum, with and without IBA annotations.} GenePT's text embedded with Qwen3-Embedding-8B, as found and with generated summaries removed; per-gene GO AUROC, the removal difference with its 95\% bootstrap interval, and the number of evaluated genes (genes with at least one benchmark term under that label set). ``Fixed set'': standard-label scores for genes evaluable under both label sets. The term-size filter is reapplied for each label set.}
  \label{tab:zerobin}
  \small
  \begin{tabular}{@{}llrcccl@{}}
    \toprule
    Probe & Labels & Bin & $n$ & As found & Removed & As found $-$ removed (95\%) \\
    \midrule
    linear & standard & 0 & 568 & 0.962 & 0.935 & $+0.027$ (+0.014 to +0.040) \\
     & IBA excluded & 0 & 100 & 0.861 & 0.772 & $+0.088$ (+0.031 to +0.145) \\
     & standard, fixed set & 0 & 90 & 0.871 & 0.780 & $+0.091$ (+0.039 to +0.144) \\
     & standard & 1-10 & 4,310 & 0.897 & 0.819 & $+0.078$ (+0.072 to +0.084) \\
     & IBA excluded & 1-10 & 3,511 & 0.877 & 0.791 & $+0.086$ (+0.079 to +0.093) \\
     & standard, fixed set & 1-10 & 3,405 & 0.883 & 0.797 & $+0.085$ (+0.078 to +0.092) \\
     & standard & 0-10 & 4,878 & 0.904 & 0.833 & $+0.072$ (+0.067 to +0.077) \\
     & IBA excluded & 0-10 & 3,611 & 0.876 & 0.790 & $+0.086$ (+0.079 to +0.093) \\
     & standard, fixed set & 0-10 & 3,495 & 0.882 & 0.797 & $+0.085$ (+0.078 to +0.093) \\
    \midrule
    mlp & standard & 0 & 568 & 0.985 & 0.973 & $+0.012$ (+0.007 to +0.018) \\
     & IBA excluded & 0 & 100 & 0.963 & 0.913 & $+0.050$ (+0.030 to +0.070) \\
     & standard, fixed set & 0 & 90 & 0.956 & 0.932 & $+0.024$ (+0.006 to +0.046) \\
     & standard & 1-10 & 4,310 & 0.952 & 0.918 & $+0.033$ (+0.030 to +0.036) \\
     & IBA excluded & 1-10 & 3,511 & 0.943 & 0.907 & $+0.036$ (+0.033 to +0.040) \\
     & standard, fixed set & 1-10 & 3,405 & 0.945 & 0.908 & $+0.037$ (+0.033 to +0.040) \\
     & standard & 0-10 & 4,878 & 0.956 & 0.925 & $+0.031$ (+0.028 to +0.034) \\
     & IBA excluded & 0-10 & 3,611 & 0.944 & 0.907 & $+0.036$ (+0.033 to +0.040) \\
     & standard, fixed set & 0-10 & 3,495 & 0.945 & 0.909 & $+0.036$ (+0.033 to +0.040) \\
    \bottomrule
  \end{tabular}
\end{table}

%% file: table_r14_cluster.tex

\begin{table}[p]
  \centering
  \caption{\textbf{Gene-level and family-clustered bootstrap intervals, and the ranking under different T1 weights.} Headline GeneProv Index (linear probe, genes with 0--10 publications) recomputed from the regenerated out-of-fold arrays of round 14 (reproduction exceptions and their treatment are documented in Methods). Gene-level intervals resample genes with replacement; family-clustered intervals resample HGNC gene groups (first group of multi-group genes; ungrouped genes are singletons; 3,018 clusters over 7,120 genes), 1,000 resamples each. Rank columns give the rank and its 95\% interval under each bootstrap at T1 weight 1/3 (the index) and the point rank at weights 0, 1/4 and 1/2, the remainder split equally between T2 and T4. The 1--10 removal difference of GenePT's text (linear) is 0.078, gene-level interval 0.072--0.084, family-clustered 0.070--0.086; MLP 0.033, 0.030--0.036 and 0.030--0.037. The last two columns hold T2's label set fixed to the headline stratum's ten labels (ClinGen excluded) on all genes: T2 and the all-genes index under that set (v1 values use all eleven labels).}
  \label{tab:cluster}
  \scriptsize\setlength{\tabcolsep}{1.7pt}
  \resizebox{\textwidth}{!}{%
  \begin{tabular}{@{}P{0.225\linewidth}cccccccccc@{}}
    \toprule
    Representation & Index & Gene CI & Family CI & Gene rank & Family rank & $w=0$ & $1/4$ & $1/2$ & T2, fixed & Index, fixed \\
    \midrule
    GB.Protein-16B $+$ rich summary-filtered text & 47.1 & 46.3--47.8 & 41.8--51.4 & 1 (1--2) & 1 (1--3) & 7 & 2 & 1 & 0.723 (0.720) & 50.6 (50.4) \\
    ProTrek-650M & 47.0 & 46.2--47.7 & 41.3--51.7 & 2 (1--2) & 2 (1--3) & 5 & 1 & 2 & 0.759 (0.749) & 49.5 (48.8) \\
    ProtST-ESM1b & 46.3 & 45.6--47.0 & 42.4--50.0 & 3 (2--3) & 3 (1--4) & 4 & 3 & 3 & 0.791 (0.773) & 49.2 (48.0) \\
    GB.Protein-16B & 45.3 & 44.5--46.0 & 40.1--49.9 & 4 (4--4) & 4 (3--4) & 3 & 4 & 6 & 0.735 (0.719) & 47.3 (46.2) \\
    ESM1b & 43.9 & 43.1--44.7 & 39.1--48.4 & 5 (5--6) & 5 (5--6) & 8 & 5 & 8 & 0.729 (0.714) & 45.4 (44.5) \\
    ESM2-650M & 43.6 & 42.9--44.4 & 38.9--48.0 & 6 (5--7) & 6 (5--8) & 9 & 6 & 10 & 0.745 (0.728) & 45.2 (44.1) \\
    GenePT 3-large (as released) & 43.0 & 42.3--43.8 & 38.6--46.9 & 7 (6--9) & 7 (5--9) & 12 & 7 & 4 & 0.713 (0.713) & 46.1 (46.0) \\
    Rich summary-filtered text & 42.8 & 42.0--43.6 & 38.7--46.4 & 8 (7--9) & 8 (6--9) & 13 & 8 & 5 & 0.723 (0.722) & 47.9 (47.8) \\
    UniProt function text only & 42.5 & 41.7--43.2 & 38.0--46.3 & 9 (8--9) & 9 (7--10) & 15 & 10 & 7 & 0.714 (0.706) & 47.1 (46.6) \\
    GenePT text as found & 41.5 & 40.7--42.2 & 37.2--45.3 & 10 (10--11) & 10 (10--11) & 16 & 12 & 9 & 0.714 (0.713) & 44.5 (44.5) \\
    GenePT ada-002 (as released) & 41.4 & 40.7--42.2 & 37.9--44.9 & 11 (10--11) & 11 (9--12) & 14 & 11 & 11 & 0.711 (0.701) & 43.2 (42.5) \\
    GB.RNA-1.6B & 39.9 & 39.2--40.6 & 35.5--43.7 & 12 (12--13) & 12 (11--14) & 6 & 9 & 13 & 0.718 (0.709) & 41.1 (40.4) \\
    Rich summary-filtered text, names masked & 39.2 & 38.4--40.1 & 36.1--42.4 & 13 (12--13) & 13 (12--13) & 11 & 14 & 12 & 0.733 (0.729) & 47.1 (46.9) \\
    Geneformer V2-316M & 36.6 & 35.8--37.3 & 34.3--39.2 & 14 (14--15) & 14 (13--16) & 1 & 13 & 18 & 0.758 (0.746) & 38.6 (37.7) \\
    scELMo (GPT-3.5-written) & 36.1 & 35.2--37.0 & 32.6--39.6 & 15 (14--16) & 15 (14--16) & 17 & 16 & 15 & 0.715 (0.707) & 42.8 (42.3) \\
    Gene symbol only & 35.6 & 34.7--36.4 & 32.0--39.1 & 16 (15--17) & 16 (14--17) & 18 & 17 & 14 & 0.676 (0.676) & 39.3 (39.3) \\
    GB.Cell-100M & 34.7 & 33.9--35.5 & 32.8--37.0 & 17 (16--18) & 17 (15--18) & 2 & 15 & 19 & 0.773 (0.757) & 38.7 (37.6) \\
    GenePT text, generated removed & 34.2 & 33.3--35.1 & 30.5--37.8 & 18 (17--18) & 18 (17--18) & 19 & 18 & 16 & 0.707 (0.708) & 40.6 (40.7) \\
    Qwen2.5-written descriptions & 31.2 & 30.4--32.1 & 27.5--34.8 & 19 (19--20) & 19 (19--20) & 20 & 20 & 17 & 0.659 (0.658) & 35.8 (35.7) \\
    GB.DNA-300M & 31.0 & 30.1--31.8 & 29.2--33.1 & 20 (19--20) & 20 (19--20) & 10 & 19 & 20 & 0.697 (0.685) & 30.1 (29.3) \\
    \bottomrule
  \end{tabular}
  }
\end{table}

%% file: table_r14_coverage.tex

\begin{table}[p]
  \centering
  \caption{\textbf{Coverage of the gene universe per representation, and T4 recall at 10.} Genes with an embedding (of 19,256) and genes without one by exposure bin (they receive the mean embedding in every task); counted separately below: genes without a summary in GenePT's 2023 text (key absent or no summary), genes without a RefSeq summary or UniProt function text in the rich input, and genes without a benchmark GO term (no T1 label). Recall at 10: share of a gene's ten DepMap co-essential partners among its ten nearest candidates by embedding cosine (genes with 0--10 publications; all genes), beside the T4 AUROC. Genes covered by every Track O row: 18,308 (6,354 with 0--10 publications); on them the headline ranking has Spearman 0.99 with the all-gene ranking and no row moves by more than 2 ranks.}
  \label{tab:coverage}
  \scriptsize\setlength{\tabcolsep}{1.7pt}
  \resizebox{\textwidth}{!}{%
  \begin{tabular}{@{}P{0.23\linewidth}rrrrrrrcccc@{}}
    \toprule
    & & & \multicolumn{5}{c}{Missing by publication count} & \multicolumn{2}{c}{0--10 papers} & \multicolumn{2}{c}{All genes} \\
    Representation & Present & Missing & 0 & 1-10 & 11-100 & 101-1000 & $>1000$ & AUROC & R@10 & AUROC & R@10 \\
    \midrule
    GB.Protein-16B & 19,133 & 123 & 16 & 59 & 42 & 6 & 0 & 0.627 & 0.017 & 0.616 & 0.016 \\
    ESM2-650M & 19,133 & 123 & 16 & 59 & 42 & 6 & 0 & 0.592 & 0.015 & 0.582 & 0.013 \\
    GB.Cell-100M & 18,593 & 663 & 295 & 277 & 71 & 19 & 1 & 0.580 & 0.005 & 0.578 & 0.008 \\
    Geneformer V2-316M & 19,229 & 27 & 4 & 11 & 10 & 2 & 0 & 0.588 & 0.004 & 0.590 & 0.004 \\
    GenePT ada-002 (as released) & 18,942 & 314 & 85 & 168 & 46 & 13 & 2 & 0.562 & 0.017 & 0.565 & 0.022 \\
    GenePT 3-large (as released) & 18,945 & 311 & 85 & 166 & 45 & 13 & 2 & 0.571 & 0.019 & 0.580 & 0.027 \\
    scELMo (GPT-3.5-written) & 19,231 & 25 & 10 & 13 & 2 & 0 & 0 & 0.559 & 0.017 & 0.572 & 0.024 \\
    Qwen2.5-written descriptions & 19,256 & 0 & 0 & 0 & 0 & 0 & 0 & 0.549 & 0.016 & 0.557 & 0.021 \\
    Rich summary-filtered text & 19,256 & 0 & 0 & 0 & 0 & 0 & 0 & 0.576 & 0.020 & 0.587 & 0.031 \\
    Rich summary-filtered text, names masked & 19,256 & 0 & 0 & 0 & 0 & 0 & 0 & 0.575 & 0.017 & 0.587 & 0.029 \\
    GenePT text as found & 19,256 & 0 & 0 & 0 & 0 & 0 & 0 & 0.565 & 0.015 & 0.571 & 0.023 \\
    GenePT text, generated removed & 19,256 & 0 & 0 & 0 & 0 & 0 & 0 & 0.556 & 0.014 & 0.553 & 0.021 \\
    Gene symbol only & 19,256 & 0 & 0 & 0 & 0 & 0 & 0 & 0.557 & 0.018 & 0.566 & 0.024 \\
    GB.DNA-300M & 19,236 & 20 & 4 & 6 & 5 & 4 & 1 & 0.582 & 0.004 & 0.571 & 0.003 \\
    GB.RNA-1.6B & 19,228 & 28 & 6 & 8 & 9 & 4 & 1 & 0.628 & 0.008 & 0.613 & 0.006 \\
    ESM1b & 19,133 & 123 & 16 & 59 & 42 & 6 & 0 & 0.612 & 0.017 & 0.601 & 0.017 \\
    ProtST-ESM1b & 19,133 & 123 & 16 & 59 & 42 & 6 & 0 & 0.574 & 0.014 & 0.570 & 0.013 \\
    ProTrek-650M & 19,133 & 123 & 16 & 59 & 42 & 6 & 0 & 0.603 & 0.017 & 0.592 & 0.016 \\
    UniProt function text only & 19,256 & 0 & 0 & 0 & 0 & 0 & 0 & 0.574 & 0.020 & 0.585 & 0.031 \\
    GB.Protein-16B $+$ rich summary-filtered text & 19,133 & 123 & 16 & 59 & 42 & 6 & 0 & 0.627 & 0.024 & 0.628 & 0.032 \\
    Rich summary-filtered text without UniProt text & 19,256 & 0 & 0 & 0 & 0 & 0 & 0 & 0.568 & 0.017 & 0.567 & 0.024 \\
    GB.Protein-16B $+$ text without UniProt & 19,133 & 123 & 16 & 59 & 42 & 6 & 0 & 0.619 & 0.021 & 0.613 & 0.027 \\
    ESM2-650M $+$ rich summary-filtered text & 19,133 & 123 & 16 & 59 & 42 & 6 & 0 & 0.600 & 0.022 & 0.607 & 0.030 \\
    ESM2-650M $+$ text without UniProt & 19,133 & 123 & 16 & 59 & 42 & 6 & 0 & 0.592 & 0.019 & 0.590 & 0.025 \\
    \midrule
    No summary in GenePT's 2023 text & 17,583 & 1,673 & 324 & 795 & 481 & 69 & 4 & & & & \\
    No RefSeq summary or UniProt text (rich input) & 17,320 & 1,936 & 654 & 1175 & 98 & 9 & 0 & & & & \\
    No benchmark GO term & 16,701 & 2,555 & 694 & 1548 & 291 & 22 & 0 & & & & \\
    \bottomrule
  \end{tabular}
  }
\end{table}

%% file: table_r14_models.tex
\clearpage
\begingroup
\scriptsize\setlength{\tabcolsep}{3pt}
\setlength{\LTcapwidth}{\textwidth}
\begin{longtable}{@{}P{0.17\textwidth}P{0.30\textwidth}P{0.10\textwidth}P{0.18\textwidth}P{0.09\textwidth}r@{}}
\caption{\textbf{Training information, relation to the evaluation labels and T3 eligibility of the representations.}
  Training data and dates are as stated by each model's authors (paper or model card); ``annotation or text signal''
  says whether any functional annotation or text about genes entered training. T3 eligibility follows the historical
  annotation-access rule and its date proxies (Methods): sequence and expression models qualify whatever their training
  date; text-containing representations qualify when the declared training-data cutoff or summary tag is pre-2023.
  These proxies do not verify that all information predates the cutoff. Eligibility is not a requirement that every model was available
  at the cutoff. Coverage is the number of the 19,256 universe genes with an embedding (the rest receive the
  mean vector; Supplementary Table~\ref{tab:coverage}).}
  \label{tab:models} \\
    \toprule
    Representation & Training data (as stated by the authors) & Data date & Annotation or text signal in training & T3 eligible & Coverage \\
    \midrule
\endfirsthead
\caption[]{Training information, label provenance and T3 eligibility (continued).} \\
    \toprule
    Representation & Training data (as stated by the authors) & Data date & Annotation or text signal in training & T3 eligible & Coverage \\
    \midrule
\endhead
\midrule
\multicolumn{6}{r@{}}{\emph{Continued on next page}} \\
\endfoot
\bottomrule
\endlastfoot
    \multicolumn{6}{@{}l}{\emph{Sequence and expression models (Track O)}} \\
    GB.Protein-16B \citep{sun2024aidoprotein} & 1.2 trillion amino acids from UniRef90 and ColabFoldDB; masked language modelling & released Nov 2024; database releases not stated & none (sequences only) & yes, by the sequence rule & 19,133 \\
    ESM2-650M \citep{lin2023esm2} & UniRef50 clusters sampled from UniRef90 (UniRef release 2021\_04), masked language modelling & 2021 & none & yes & 19,133 \\
    ESM1b \citep{rives2021esm1b} & UniRef50 2018\_03, masked language modelling & 2018 & none & yes & 19,133 \\
    GB.Cell-100M \citep{ho2024aidocell} & 50 million single-cell transcriptomes from over 100 tissue types, read-depth-aware masked modelling & released Nov 2024 & none (expression counts) & yes, by the expression rule & 18,593 \\
    Geneformer V2-316M \citep{chen2024geneformerv2} & about 104 million human single-cell transcriptomes (Genecorpus-104M), rank-value encoding & trained Dec 2024 & none & yes & 19,229 \\
    GB.RNA-1.6B & 42 million non-coding RNA sequences, RNAcentral 24.0 & 2024 & none & yes & 19,228 \\
    GB.DNA-300M & genomes of 796 species (10.6 billion nucleotides; NCBI assemblies) & 2024 & none & yes & 19,236 \\
    \multicolumn{6}{@{}l}{\emph{Text-aligned protein encoders (Track O; embed no text)}} \\
    ProtST-ESM1b \citep{xu2023protst} & ESM1b aligned with Swiss-Prot function text (ProtDescribe) & Swiss-Prot 2022 & UniProt function text (co-curated with GOA, Section~\ref{sec:prov}) & yes (text predates the cutoff) & 19,133 \\
    ProTrek-650M \citep{su2024protrek} & ESM2-650M aligned with Swiss-Prot text and structure, contrastive & Swiss-Prot 2024 & UniProt function text & no (text after the cutoff) & 19,133 \\
    \multicolumn{6}{@{}l}{\emph{Text representations (Track O; text and encoder dates both matter)}} \\
    GenePT ada-002, as released \citep{chen2024genept} & NCBI Gene summaries (October 2023 snapshot) embedded with OpenAI text-embedding-ada-002 & text Oct 2023; encoder 2022 & 31\% of the embedded summaries are generated from GO & yes for the evaluation genes: their summaries carry tags dated before 2023 (tag-date rule, Methods) & 18,942 \\
    GenePT text-embedding-3-large, as released & the same NCBI text plus UniProt summaries, text-embedding-3-large & text 2024; encoder 2024 & generated summaries and UniProt text & no & 18,945 \\
    scELMo, GPT-3.5-written \citep{liu2023scelmo} & descriptions written by GPT-3.5 from its own memory, embedded with ada-002 & reported GPT-3.5 training-data cutoff 2021 & LLM memory of the literature and databases & yes, by the reported training-cutoff proxy & 19,231 \\
    Qwen2.5-7B-written descriptions & descriptions written by Qwen2.5-7B-Instruct \citep{qwen2024qwen25}, embedded with Qwen3-Embedding-8B & LLM 2024; encoder 2025 & LLM memory & no & 19,256 \\
    Rich summary-filtered text; names masked; UniProt text only; symbol only (Qwen3-Embedding-8B) & NCBI RefSeq summaries and UniProt function text of October 2026, or the symbol alone & text 2026; encoder 2025 & UniProt text (co-curated); encoder memory may hold NCBI pages newer than the cutoff & no & 19,256 \\
    GenePT text as found; generated summaries removed (Qwen3-Embedding-8B) & GenePT's 2023 NCBI text, re-encoded & text Oct 2023; encoder 2025 & generated summaries (as found) & no (encoder after the cutoff) & 19,256 \\
    GB.Protein-16B $\oplus$ rich summary-filtered text & concatenation of two rows above & mixed & UniProt text & no & 19,256 \\
    \multicolumn{6}{@{}l}{\emph{Track F embedders (scored on GenePT's 2023 text with generated summaries removed)}} \\
    Qwen3-Embedding-8B; -0.6B \citep{zhang2025qwen3embedding} & Qwen3 language models tuned for embedding on web-scale text pairs & 2025 & LLM memory & -- & -- \\
    all-mpnet-base-v2 \citep{song2020mpnet,reimers2019sbert} & MPNet fine-tuned on one billion sentence pairs & 2021 (base model 2020) & general text & -- & -- \\
    bge-large-en-v1.5 \citep{xiao2023cpack} & BERT-large trained on web text pairs & Sep 2023 & general text & -- & -- \\
    pubmedbert-base-embeddings \citep{gu2021pubmedbert} & PubMedBERT (PubMed abstracts and full text) fine-tuned for sentence embeddings & 2023 (base model 2020) & biomedical literature & -- & -- \\
    TF-IDF, 1,024-d SVD & fitted on the two corpora themselves & -- & none & -- & -- \\
\end{longtable}
\endgroup

%% file: table_r14_rounds.tex
\begin{table}[p]
  \centering
  \caption{\textbf{Record of the analysis rounds.} Plan written: the sign-off time recorded inside the plan file;
  plan and results committed: the commit times of the plan file and of the round's results in the repository (the
  first eight characters of the commit hash). ``Added after the plan'' lists analyses that were not in the plan and
  are marked post hoc in the text. Rounds 4--6 were exploratory rounds whose results are not used here beyond the
  round-4 text embeddings and label files; round 4 committed its plan (5b3cdcb, 06:03) before its results (1da05b4,
  06:54), rounds 5b and 6 committed plan and results together; their gate logs are in their plan files.}
  \label{tab:rounds}
  \scriptsize\setlength{\tabcolsep}{2pt}
  \begin{tabular}{@{}P{0.05\textwidth}P{0.19\textwidth}P{0.11\textwidth}P{0.12\textwidth}P{0.12\textwidth}P{0.20\textwidth}P{0.15\textwidth}@{}}
    \toprule
    Round & Content & Plan written (UTC) & Plan committed & Results committed & Added after the plan (post hoc) & Gates \\
    \midrule
    7 & the circularity audit under a documented analysis plan: prevalence, inflation, label sets, encoders, published embeddings and benchmarks, temporal holdouts, time travel, assay labels (Sections~\ref{sec:prev}--\ref{sec:published}, \ref{sec:temporal}, \ref{sec:know}) & 2026-10-09 03:57 (signed off by the author) & 2026-10-09 15:38, 561c9c2, with the results & same commit & protein model against frozen literature-only text; fixed-model assay comparison; text-versus-protein by exposure; GenePT's 2023 text in the Zhong et al.\ temporal task (prediction wrong, logged) & 16 passed; 4 failed (D3: non-GO labels move; G1, G2: time-travel and fusion thresholds; S for ProtST) \\
    8 & LLM-written descriptions, scELMo, model organisms, UniProt statements inferred from sequence (Sections~\ref{sec:llm}, \ref{sec:organisms}, \ref{sec:know}) & 2026-10-10 01:02 & 2026-10-10 01:43, 41d7aeb, with the results & same commit & protein model against LLM-written text by exposure; ESM2 in place of GB.Protein-16B; correction of the verbatim-overlap figures (summary-only text) & 7 passed; 1 failed (H2: time travel against frozen LLM text) \\
    9 & simulated review and revision; UniProt text in place of the summaries, nearest-neighbour probe, fold seeds, family-restricted time travel (Section~\ref{sec:prov}) & 2026-10-10, before 05:19 & 2026-10-10 05:19, 3790283, before computation & 2026-10-10 15:12, df9542c & co-curation test (declared before it was computed); replacement by LLM text; per-term AUROC within bins; ESM2 on the zero-literature comparison; one detector phrase added after the planned evaluation & 3 passed (L, M, N); 2 failed (K1, K2: UniProt text recovers the loss) \\
    10 & GeneProv v0, the frozen benchmark table & 2026-10-10, before 15:37 & 2026-10-10 15:37, e4ebbd7, before computation & 2026-10-10 16:02, 8026128 & none & 4 passed (R, V1--V3) \\
    11 & GeneProv v1: the index, seven rows, Track F index, leaderboard & 2026-10-11, before 07:25 & 2026-10-11 07:25, e541de0, before computation & 2026-10-11 08:10, 779a1d5 & none; the first run of the table script applied gate I1 to the text-aligned encoders, contrary to the plan's wording, and was corrected and rerun (logged 07:53, before the rerun) & 4 passed (R, I1--I3); predictions without gates held \\
    12, 13 & restructuring of the manuscript into two parts; close-out (author matter, deposit) & 2026-10-12 & 2026-10-12 06:40, 3a99920; 18:43, e67c476 & 06:52, 8052098; 18:48, d94d355 & no number computed & none \\
    14 & response to the external comments: appended rows, clustered bootstrap, weight and coverage sensitivity, baselines (this version) & 2026-10-17, before 16:57 & 2026-10-17 16:57, cc3c0ea; pushed to GitHub 16:57:10, branch round-14 (9b1c678), before computation & 2026-10-17, with this version (hash in the repository log) & none; every analysis was in the plan, and the four predictions that failed (P3 in part, P4 in part, P9 in part, P11) are reported as such & reproduction gate: 350 of 366 cells within tolerance; the 16 failing cells (ESM2-650M re-embedding; four MLP T2 cells) listed in Methods; v1 values unchanged \\
    \bottomrule
  \end{tabular}
\end{table}